\RequirePackage{fix-cm}
\documentclass[twocolumn]{svjour3}          
\smartqed  
\usepackage{hyperref}

\usepackage{graphicx}
\usepackage{subfig}
\usepackage{natbib}
\usepackage{makecell}
\usepackage{array}
\usepackage{amssymb}
\usepackage{multirow}

\usepackage{overpic}
\usepackage{color}

\usepackage{booktabs}
\usepackage[table,xcdraw]{xcolor}

\makeatletter  
\newif\if@restonecol  
\makeatother

\usepackage[linesnumbered,ruled,vlined]{algorithm2e}
\usepackage{algpseudocode}  
\usepackage{amsmath}  

\newcommand{\markCM}{\textcolor{black}}

\newcommand{\noteCM}{\textcolor{blue}}

\begin{document}
	
	\title{Hierarchical Topometric Representation of 3D Robotic Maps
	}
	
	
	\author{Zhenpeng He \and	
		Hao Sun \and Jiawei Hou \and Yajun Ha \and S\"{o}ren Schwertfeger 
	}
	
	\institute{Zhenpeng He, Hao Sun, Jiawei Hou, Yajun Ha, and S\"oren Schwertfeger are with the School of Information Science and Technology,	ShanghaiTech University, Shanghai 201210, China.
		\email{\{hezhp, sunhao, houjw, hayj, soerensch\}\\@shanghaitech.edu.cn} 
	}

	\maketitle

\begin{abstract}
	In this paper, we propose a method for generating a hierarchical, volumetric topological map from 3D point clouds. There are three basic hierarchical levels in our map: $storey - region - volume$. The advantages of our method are reflected in both input and output. In terms of input, we accept multi-storey point clouds and building structures with sloping roofs or ceilings. In terms of output, we can generate results with metric information of different dimensionality, that are suitable for different robotics applications. The algorithm generates the volumetric representation by generating $volumes$ from a 3D voxel occupancy map. We then add $passage$s (connections between $volumes$), combine small $volumes$ into a big $region$ and use a 2D segmentation method for better topological representation. We evaluate our method on several freely available datasets. The experiments highlight the advantages of our approach.
	
\end{abstract}

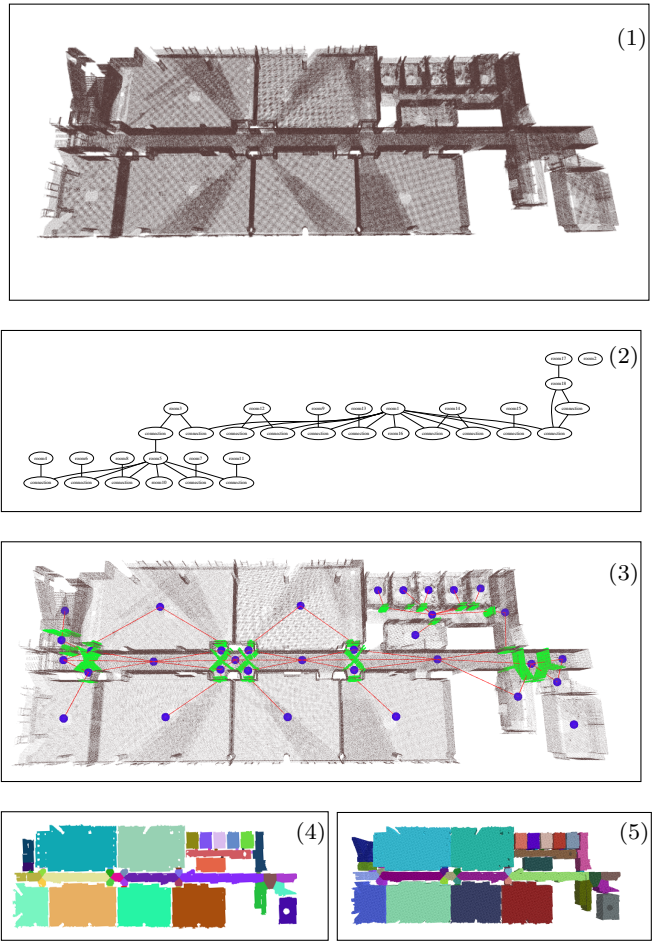
\begin{figure}[ht!]
	\centering
	\subfloat{
		\label{fig:p1_1}
		\fbox{\includegraphics[width=0.95\linewidth]{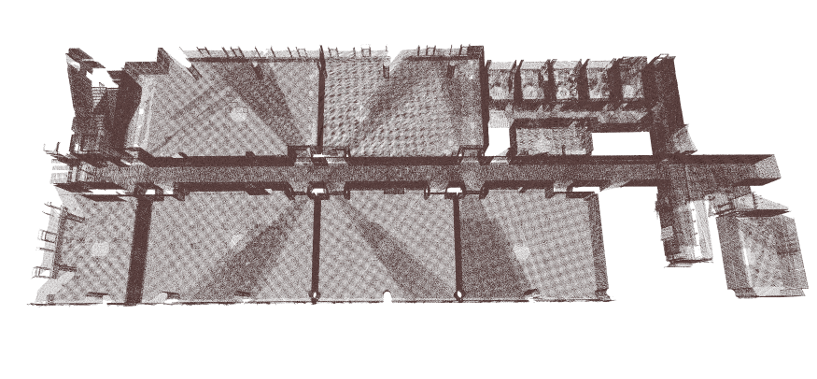}
              \begin{picture}(0,0)
		          \put(-6,93){(1)}
              \end{picture}
	         }
	} \\ 		\subfloat{
		\label{fig:p1_2}
		\fbox{\includegraphics[width=0.95\linewidth]{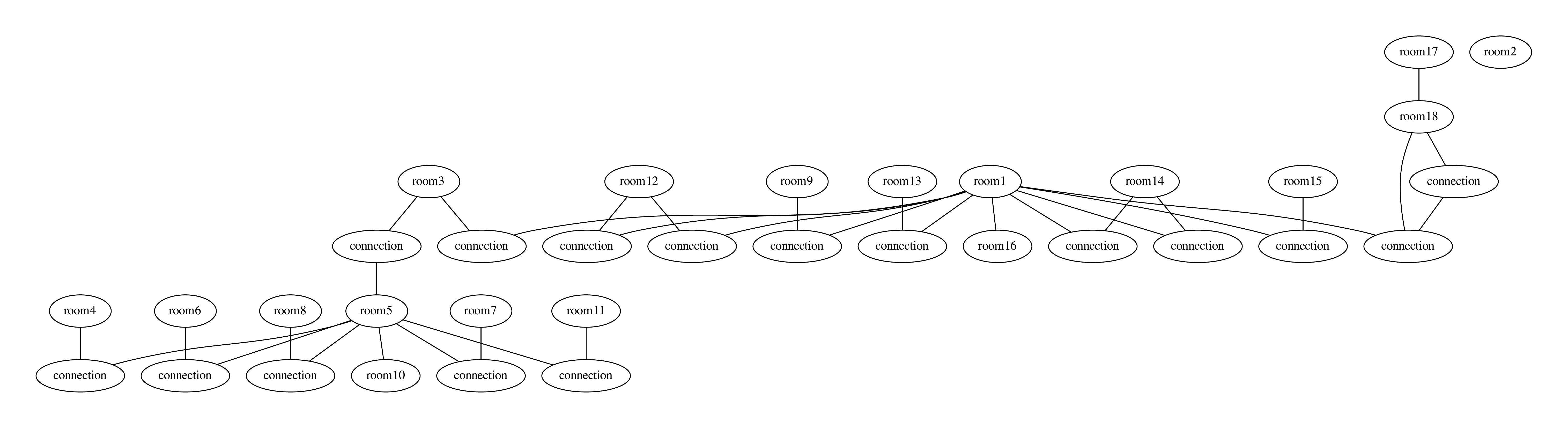}
              \begin{picture}(0,0)
				 \put(-6,53){(2)}
			  \end{picture}	
			 }
	} \\		
	\subfloat{
		\label{fig:p1_3}
		\fbox{\includegraphics[width=0.95\linewidth]{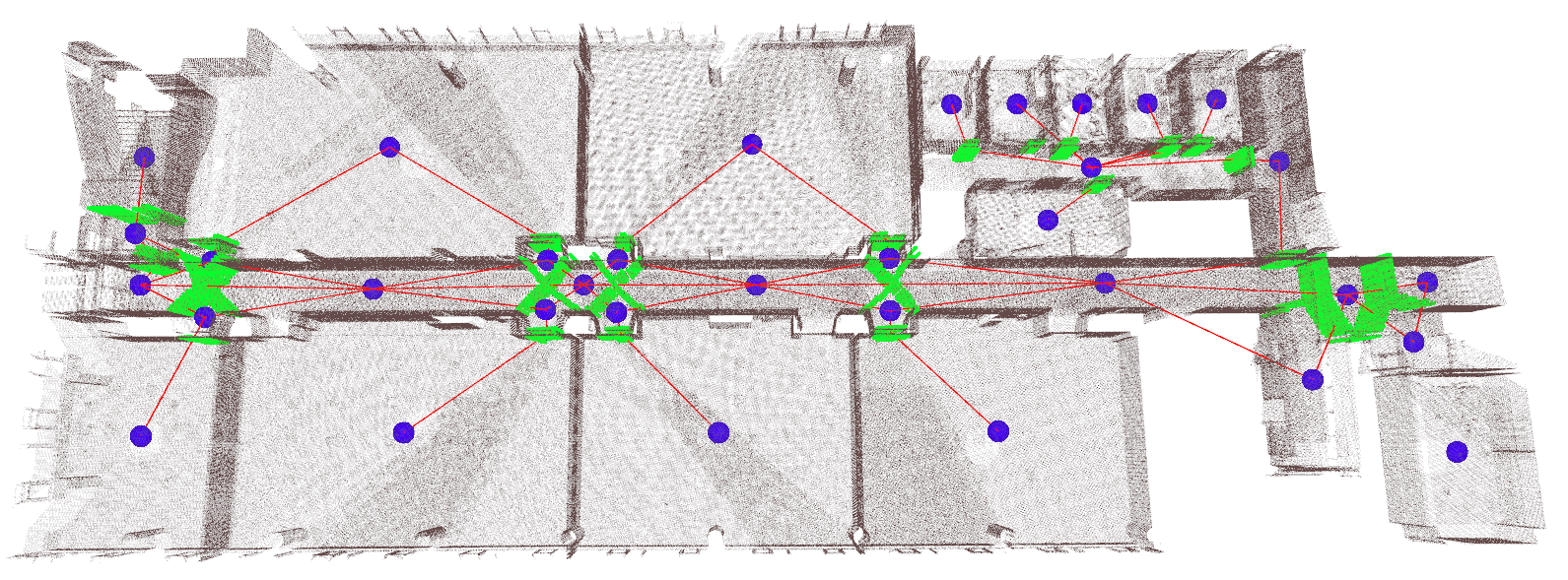}
              \begin{picture}(0,0)
					\put(-6,73){(3)}
			  \end{picture}			
	}
	} \\	
	\subfloat{
		\label{fig:p1_4}
		\fbox{\includegraphics[width=0.46\linewidth]{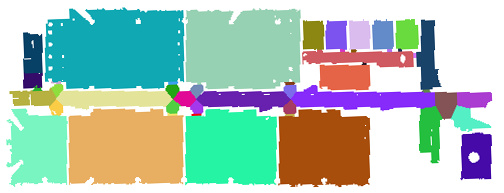}
		       \begin{picture}(0,0)
				    \put(-6,35){(4)}
		       \end{picture}	
	         }
		\fbox{\includegraphics[width=0.44\linewidth]{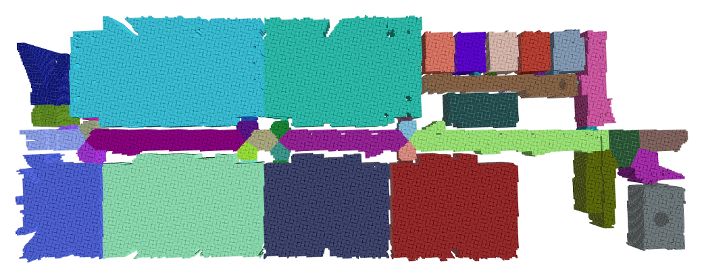}
		      \begin{picture}(0,0)
				    \put(-6,35){(5)}
		      \end{picture}	
		     }
	}				
	\caption{Input 3D point cloud (1), and the different outputs provided by our method: 0D topological map (2); 1D map with coordinates annotated to the vertices (3), 2D map with colored regions (4); and 3D map with volumetric information (5).}
	\label{fig:p1}
\end{figure} 	
     
\section{Introduction}
\label{sec:intro}

Topometric maps are generated on top of metric maps,  by partitioning the latter into coherent regions as topological vertices. It combines advantages of both metric maps and topological maps. For navigation, topological maps can be queried for very fast initial global plans, and metric maps can further refine the local plans \citet{oleynikovasparse}. Topometric localization combines topological and metric localization to achieve the reliability of topological localization with the geometric accuracy of metric localization \citet{badino2012real}. 

Extracting the topological structure of a 2D grid map has been a well-studied topic \citet{thrun1998learning, kuipers2004local}. The topometric map based on 3D space is a recent research \citet{blochliger2018topomap}. There are multiple challenges to extend this representation to 3D space: 1) In indoor environments, objects like furniture frequently occupy free space, making it hard to partition. 2) The storage space of 3D maps often grows quadratically, but the resources of robots are limited. The most common topological map generation algorithms are those based on Voronoi diagrams. \citet{okorn2010toward} uses critical points along with Voronoi Graphs for segmentation in indoor environments. Other, often similar segmentation algorithms have been presented \citet{wurm2008coordinated, schwertfeger2015Map, schwertfeger2013evaluation} to improve the accuracy of map segmentation. Many segmentation and geometric structure extraction methods are proposed \citet{babacan2016towards, ochmann2019automatic} for 3D point clouds. However, only a few of them concentrate on topology extraction. 



\markCM{ In this paper, we propose a 3D hierarchical, volumetric topometric map representation. Inspired by $ $ \citet{blochliger2018topomap}, instead of representing obstacles, we use traversable free 3D space to represent our map. The representation supports varying dimensionality (see Fig.~ \ref{fig:p1}), a concept similar to the Spatial Semantic Hierarchy of \citet{kuipers2000spatial}. All dimensionality levels are embedded in 3D space, the dimensionality refers to the information annotating the vertices and edges of the graph. At 0D the map is a purely topological representation of free 3D spaces ($volumes$) and their connections. At 1D each vertex is annotated with a point in 3D coordinates. At 2D vertices are additionally annotated with 2D areas and edges are annotated with poly-line passages, defining the border gateway where one can traverse from one area to the next. In 3D vertices are annotated with 3D volumes and the edges are annotated with passages represented as 2D surfaces. Through those passage surfaces one can move from one $volume$ to its neighbor. }

\markCM{ Note that in the 1D representation often edges are annotated with metric path information (e.g. for 2D maps in \citet{thrun1998learning, kuipers2004local, schwertfeger2015Map}). We did not generate nor annotate such paths between the two vertices of an edge, but that could be easily done via a 3D planning algorithm (e.g. A*) on a 3D grid map generated from the input point cloud. }

\markCM{ Additionally to the dimensionality, our map representation is also hierarchical. That means that we have different levels of granularity, where a higher-level vertex is often the parent of several lower level vertices. At the lowest level we have $volumes$, encompassing free space. The next level in the hierarchy is then a $region$, often containing several $volumes$. We aim to represent each room as a $region$, but also corridors and the free space beneath doors. In fact, at this level, most regions will be connected by the small regions beneath the doors, that we label ``connection''.  Since some $regions$ may be very big, we additionally introduce a second region level that sub-divides said big $regions$.  We furthermore support input point clouds of buildings with multiple storeys, where each floor is then represented in one $storey$. Thus the basic hierarchical levels in our map are: {\it storey - region1 - region2 - volume}. }



\markCM{Our representation is generated with a 3D point cloud as input. Given a point cloud of the whole building, the $storey$ can be distinguished by a space divider. Then, on each $storey$, we take the following steps to represent the environment, which are illustrated in Fig. \ref{fig:column}. We start by sampling the point cloud of a $storey$ in a 3D grid with 3D voxels of a certain resolution. We then find $columns$, which are maximal contiguous vertically-oriented sequences of free voxels. We store a $column$ only using four parameters $(x,y,z_1,z_2)$. We  then apply the height-based merge method to combine $columns$ with a similar height to $regions$ and then generate the passage contact surfaces between neighboring $regions$.  In the last step, we segment big $regions$ in the horizontal direction by utilizing a 2D segmentation method we proposed in \citet{hou2019area}.}


\markCM{ Topometric representations, such as the one introduced in this paper, can be used for various applications in robotics. Navigation and planning can be done much faster on the topometric graph instead of a grid map, \citet{thrun1998learning}. This is very interesting for global planning, where big distances can be planned for very quickly, \citet{hou2018topological, oleynikovasparse}. Hierarchical maps speed up this planning even further, as they allow to first generate a very coarse high-level roadmap before planning paths at a higher resolution, \citet{park2018incremental}. Different dimensionalities offer different options for planning. At 0D we can say if a path exists, while at 1D, given the vertex coordinates, we can also get an estimate for its metric distance/ cost. The graph annotated with 2D information (areas and passages) can be used to exactly determine the vertex in the graph. Given coordinates for an initial position and a goal position, one can generate more realistic 2D paths for ground robots between passages of the areas. The 2D areas also implicitly encode the location of obstacles, as the edges of polygons bounding areas that are not passages typically coincide with walls. The 3D volumes have the same benefits as the 2D version, but they support also the planning for robots capable of motion in all three spacial dimensions, e.g. micro aerial vehicles. }

\markCM{ Another big benefit of topological mapping in Simultaneous Localization and Mapping (SLAM) is, that it can serve as a means for validating and invalidating loop closure, \citet{savelli2004loop}, and, more generally, for correcting the drift of position/pose estimators, \citet{badino2012real}. }

\markCM{ The 2D and 3D representation also serves as a segmentation of the point cloud into specific regions, such as rooms, corridors or storeys, \citet{armeni20163d}. In so far this representation is also very useful for semantic mapping, where labels are assigned to certain regions (e.g. ``kitchen'', ``student office'', etc.) \citet{chang2017matterport3d}, facilitating human-robot interaction, \citet{khandelwal2017bwibots}.}



We evaluate our work with experiments on a variety of datasets. We also show quantitative comparisons of our 3D $region$ segmentation maps with ones produced by existing methods and with ground truth information. The main contributions of this paper are as follows:

\begin{itemize}
	\item Hierarchical topometric and volumetric map representation. 
	\item Enriched, multi-dimensional, annotations of the topological map for different applications.
	\item The first system capable of processing fully 3D point clouds with slanted ceilings.
	\item Experimental evaluation of our approach.
\end{itemize}
The rest of the paper is structured as follows: Section \ref{sec:rela} presents related work while Section \ref{sec:method} introduces the algorithm to extract the volumetric information and create the hierarchical topometric map. Section \ref{sec:ex} analyzes the performance of the proposed algorithm in several experiments. Conclusions are given in Section \ref{sec:conclusion}.

\section{Related Work}
\label{sec:rela}

Topometric maps are typically generated on top of metric maps. There are mainly two parts in topometric map generation: space parsing and topology extraction. We provide an overview of the related literature below. The following points make our approach different from existing works:  
\markCM{
	\begin{itemize}
		\item Dealing with a complete point cloud without pose information (except the direction of the $z$-$axis$, e.g. from the gravity vector of an IMU).
		\item Dealing with real indoor environments and their occlusion and clutter.
		\item A simplified description of free space, to help with robot navigation.
		\item Partitioning the free space in a manner consistent with human semantics.
	\end{itemize}
}

\subsection{Topology extraction}
There are mainly three different ways to extract topological structures from the metric map. 

$Based\ on\ Morphology:$ Voronoi diagrams are used to divide grid maps into several subregions, which have been used to construct the topological structure in grid maps, \citet{thrun1998learning}. \citet{oleynikovasparse} use an Euclidean Signed Distance Field to extract a 3D Generalized Voronoi Diagram (GVD) and obtain a thin skeleton diagram representing the topological structure of the environment. Although a generalized Voronoi graph can be used to extract topological structure, the Voronoi graph construction method is mainly suited for corridor-like environments, \citet{vazquez2015reaction}. Some of the works use the convex hull method in space into divide space. \citet{park2018incremental} incrementally generate the hierarchical roadmap (HRM) in 2D space. They decompose space to squares and then use graph cut to segment areas. HRM has a multi‐layered graphical structure that enables it to cover navigable areas using a smaller number of vertices and edges. The research of \citet{blochliger2018topomap} inspired our work. They propose an algorithm that employs, based on a volumetric occupancy grid, voxel cluster growing and merging to generate convex free space clusters from noisy and partly incomplete visual SLAM data. The convex free space clusters are the vertices in their topological map, each corresponding to a specific partially enclosed area within the environment. \markCM{ The Freespace-based map can be directly used for 3D navigation of the robot without collision with mapped obstacles. However, vertices generated based on the convex hull are hard to understand for human beings, complicating human-robot interaction. }


$Based\ on\ Semantics:$ The seminal work of \citet{kuipers2000spatial} introduced Spatial Semantic Hierarchy (SSH). The SSH has multiple levels and expresses states of partial knowledge. So it can be used for human cognitive map, robot exploration and map-building, all at the same time. In \citet{kuipers2004local} a hybrid mapping method to extend SSH is proposed. They use travel experience to abstract a local metric map to a topological place. The hybrid method combines both advantages of the metric map and the topological map to solve multiple nested large-scale loops. Learning-based methods achieve good performance in semantic map generation, \citet{hiller2020learning}. Until now, there are still few papers for 3D topometric map generation. One of the main reasons for this may be that often a 2D map is good enough for indoor navigation. Some work on element detection uses the rooms and doors as vertices to generate a topological relationship. \citet{ochmann2014towards} generate a hierarchical graph-based building topological structure based on the door and rooms. \markCM{ It achieves room segmentation by using co-visibility, but need the correspondence between scan and room as input (e.g. one scan per room). }

$Based\ on\ Distance:$ \markCM{ There is a strategy that generates topological vertices by distance. In other words, it splits the map at fixed intervals into sub-maps and creates a vertex for each sub-map. Two vertices have an edge when there is a real-world path between them. The work of \citet{badino2012real} uses these subdivided vertices to estimate the robot localization. \citet{schmuck2016hybrid} use the topological map to speed up the process of reconstruction after optimization and path planning in SLAM. Their structure is based on the metric information on a coarse scale. Both methods do not take the spatial semantic information into consideration, limiting the usefulness of this strategy in the context of human cognitive maps.}

\begin{figure*}[t]
	\centering
	\subfloat{
		\fbox{
			\includegraphics[height=5cm]{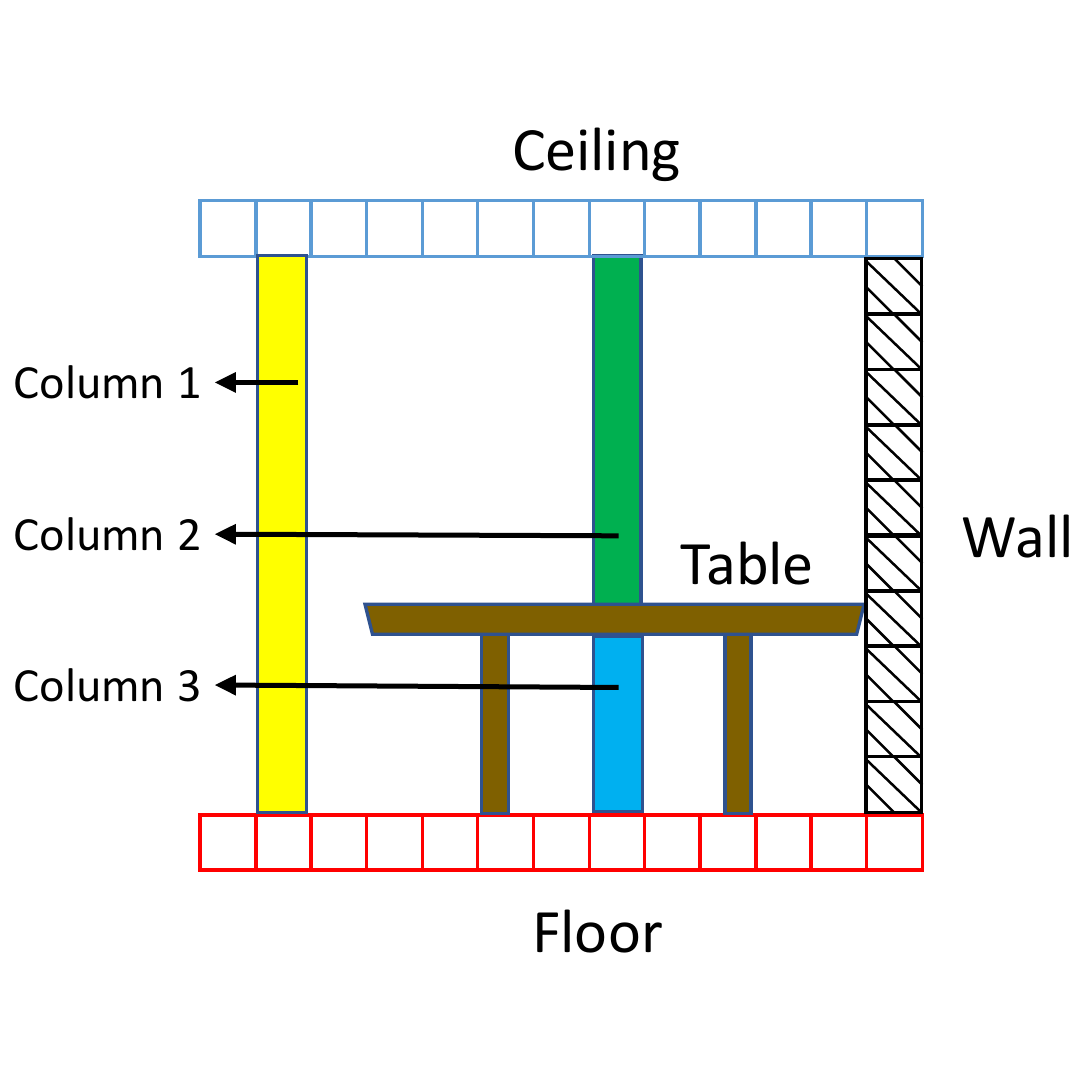} 
		}
		\fbox{
			\includegraphics[height=5cm]{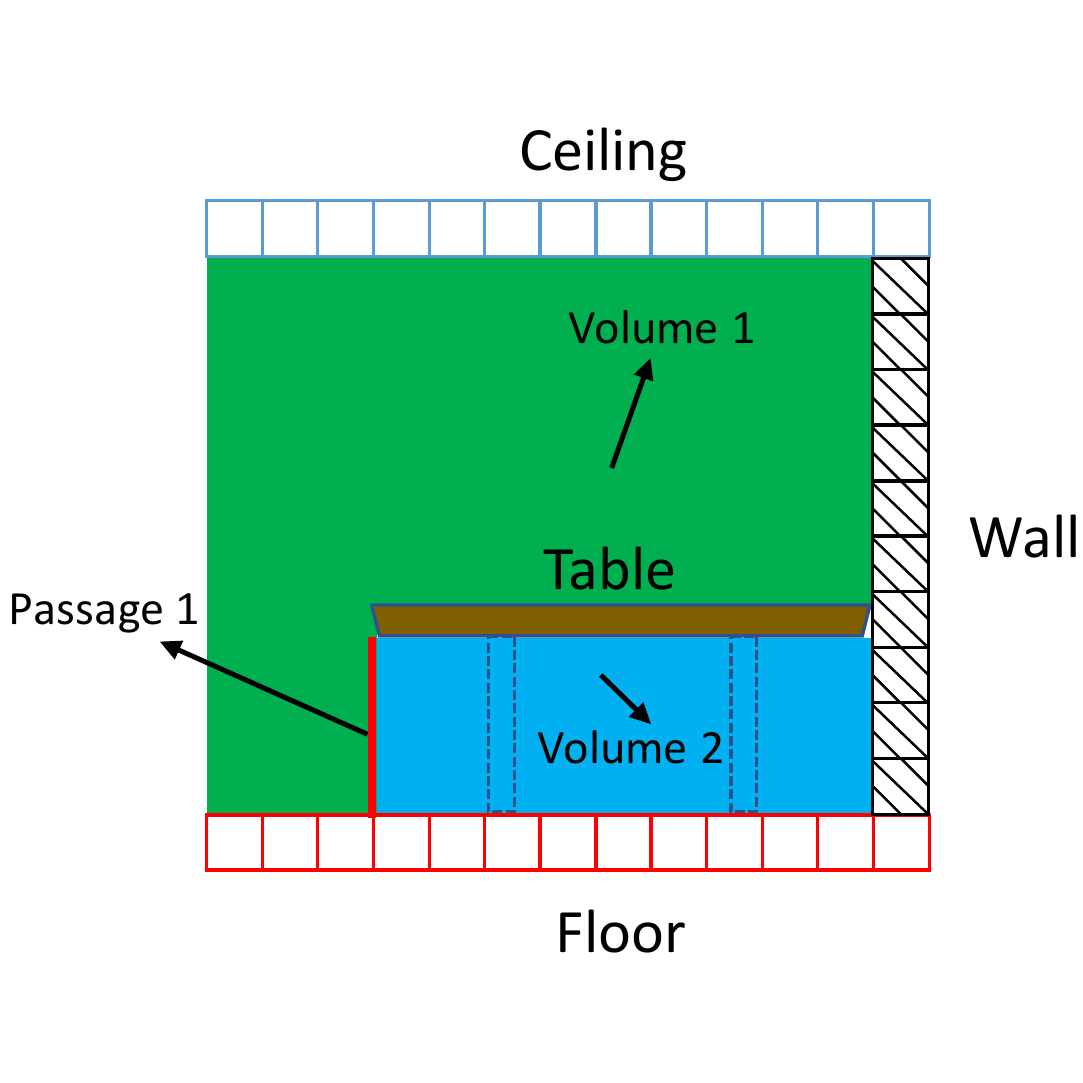} 
		}
		\fbox{
			\includegraphics[height=5cm]{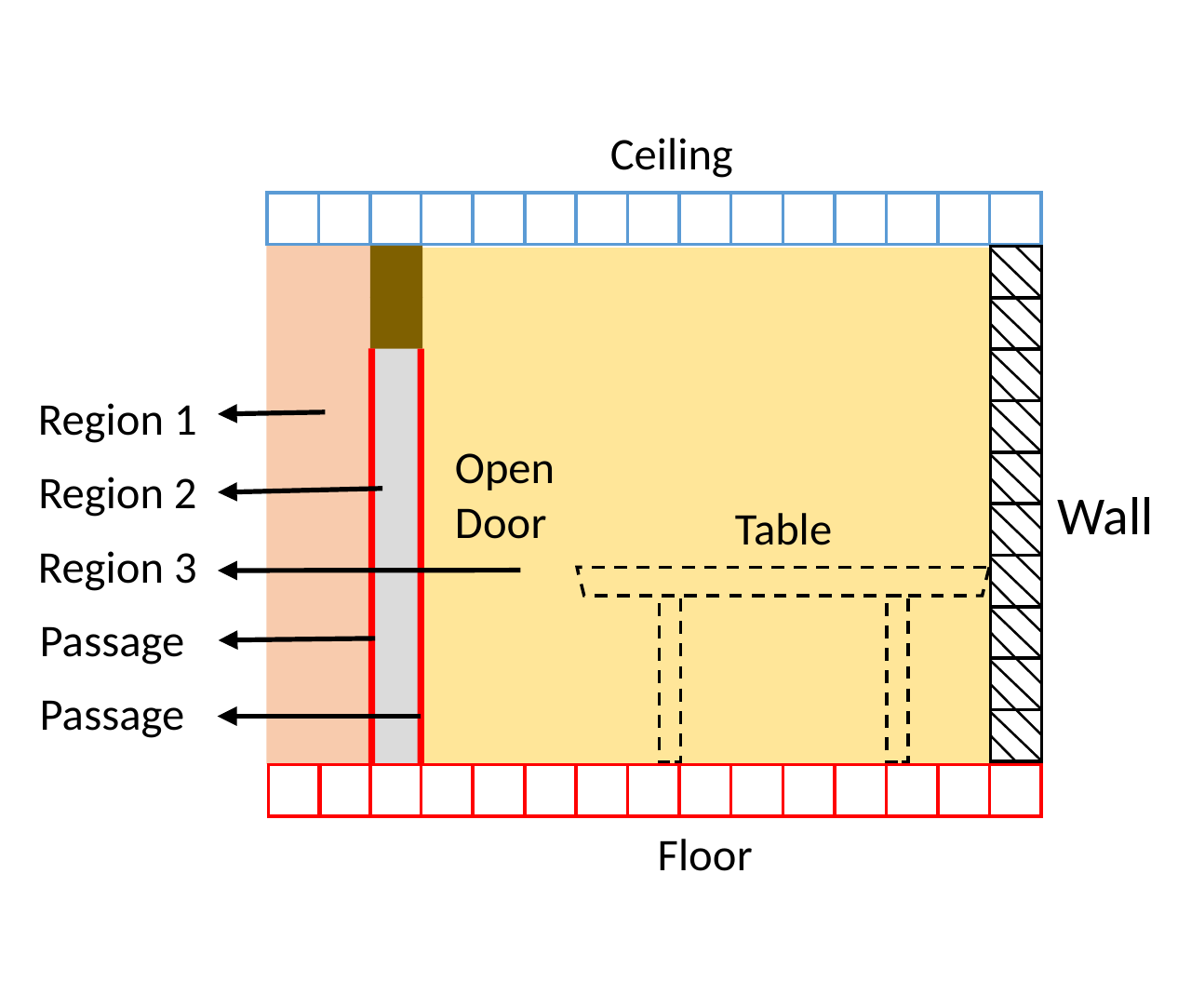}
		}
	} 		 
	
	\caption{\textbf{Left:} $columns$ generation schematic side-view diagram in a hand-made indoor environment. Only three example columns of free space are shown. \textbf{Middle:} $volumes$ and $passage$ schematic diagram. Columns with similar top height are merged into one $volume$. The $volume$ underneath the table is enclosing the legs in 3D. The $passage$ is the 2D contact surface between different $volumes$, represented as a mesh. \textbf{Right:} The example is extended to include an open door and the wall above the door. There are three $regions$: The room with Region 3, which contains Volume 1 and 2; Region 2 which has one $volume$ consisting of all the columns beneath the door; and Region 3, the room on the other side of the door. There are two $passages$ connecting the $regions$ and their $volumes$.}
	\label{fig:column}
\end{figure*} 

\subsection{Space Partitioning}

In the context of space partitioning in the building, \citet{turner2014floor} triangulated 2D wall points on a 2D planar map. The vertices of the triangulation were labelled as interior and exterior. Exterior vertices correspond to points outside the building, and they are excluded. In the next step, they use the Delaunay property of the triangulation to identify likely seed triangle locations for room labels, and then graph-cut is used to partitioning room labels. Their algorithm can exclude outliers, but still encounter the problem of clutter environment caused by furniture.  

To address the problem of occlusion and clutter, \cite{adan20113d} uses a histogram of the point cloud to identify modelled candidate surfaces. First, the walls, ceiling, and floor are detected using projections, followed by a Hough transform. Then a learning-based opening detector is used to determine holes on the surface. However, this method requires the scan data to include pose as input, which is unavailable in most of the open-source datasets. \citet{babacan2016towards} get the 2D floor plan by slicing the 3D point cloud at a height slightly below the ceiling, to avoid the influence of most furniture items. Then it uses door detection by projecting points to the 2D plane and wall plane detection with RANSAC.  The door and wall surface are detected in the process of space partitioning. Typically it can avoid most of the influence of furniture, but it assumes that the height of the ceiling is uniform, which may not be true in some of the cases. At the same time, the influence of the furniture may not totally be removed in some of the cases. \citet{armeni20163d} first finds the space dividers (e.g. walls) by a density histogram. They use a bank of peak-gap-peak filters and perform the matching operation to detect the space between walls. In the next step they are utilizing a semantic element detection in which the geometric priors are acquired from parsing into disjoint spaces. Then they reincorporate the detected elements by updating the found spaces. The approach of their work can get rid of the influence of the cluttered environment and achieves fully automatic processing of 3D point clouds. Although it is not designed for robot maps, it can be convenient to extract topological structures from the detected elements. But it assumes buildings with roughly planar walls, and hence, does not handle circular and oval-shaped rooms. \citet{ochmann2019automatic} uses an efficient RANSAC method implementation to detect planes in the building. The detected planes are used to determine point clusters corresponding to individual rooms by a ray casting approach.

\section{Method}
\label{sec:method}
In this part, we present the details of our algorithm. In Section \ref{method:representation}, an overview of the 3D hierarchical topological map representation is given. \markCM{ Section \ref{method:storey} explains how we calculate the position of the floor as well as filter the input point cloud. In Section \ref{method:column}, we show how $columns$ are generated from the 3D voxel occupancy map. Then in Section \ref{method:volume}, $volumes$ are assembled from contiguous $columus$. Afterwards we use pseudocode to explain the $passage$ generation in pipeline form. } Next, $regions$ are obtained in Section \ref{method:region} by the topological relationship between $volumes$. In Section \ref{method:area}, we perform 2D Area Graph segmentation based on the 2D grid map, which is derived from the $region$. Furthermore, we can generate a complete topometric map by combining 2D segmentation results.	    

\subsection{Topometric Map Representation}
\label{method:representation}
In this section, we give an overview of the hierarchical topometric representation in our work. The three levels of the topological graph are $storey - region - volume$.  The input of our algorithm is a complete building's point cloud. We have three kinds of vertices, $storey$, $region$, $volume$. The correspondence between elements in the topological graph and the real world is shown in Table \ref{table:correspondence}. In the $storey$ level, the vertex is one floor of the building. The $storey$ contains multiple $regions$, which abstract the enclosed space (e.g. room). Similarly, one $region$ contains multiple $volumes$, which correspond to the free space. \markCM{Two vertices are joined by an edge when they share a $passage$ (contact surface). A schematic diagram of the vertices and $passages$ can also be found in Fig. \ref{fig:column}.} 
    
    \begin{table}[h!]
    	\begin{tabular}{|l|l|c|ll}
    		\cline{1-3}
    		\textbf{Level}  & \textbf{Vertex}  & \multicolumn{1}{l|}{\textbf{Edge}}                                                                  &  &  \\ \cline{1-3}
    		$storey$ & floor of the building & \multirow{3}{*}{\begin{tabular}[c]{@{}c@{}}  $passage$ \\ connecting vertices, \\ traversable \end{tabular}} &  &  \\ \cline{1-2}
    		$region$ & enclosed space        &                                                                                            &  &  \\ \cline{1-2}
    		$volume$ & free space cluster    &                                                                                            &  &  \\ \cline{1-3}
    	\end{tabular}
     	\caption{The correspondence between elements in the topological graph and the real world.}
    	\label{table:correspondence}
    \end{table}

\begin{figure}[htb]
	\subfloat{
		\fbox{\includegraphics[width=0.9\linewidth]{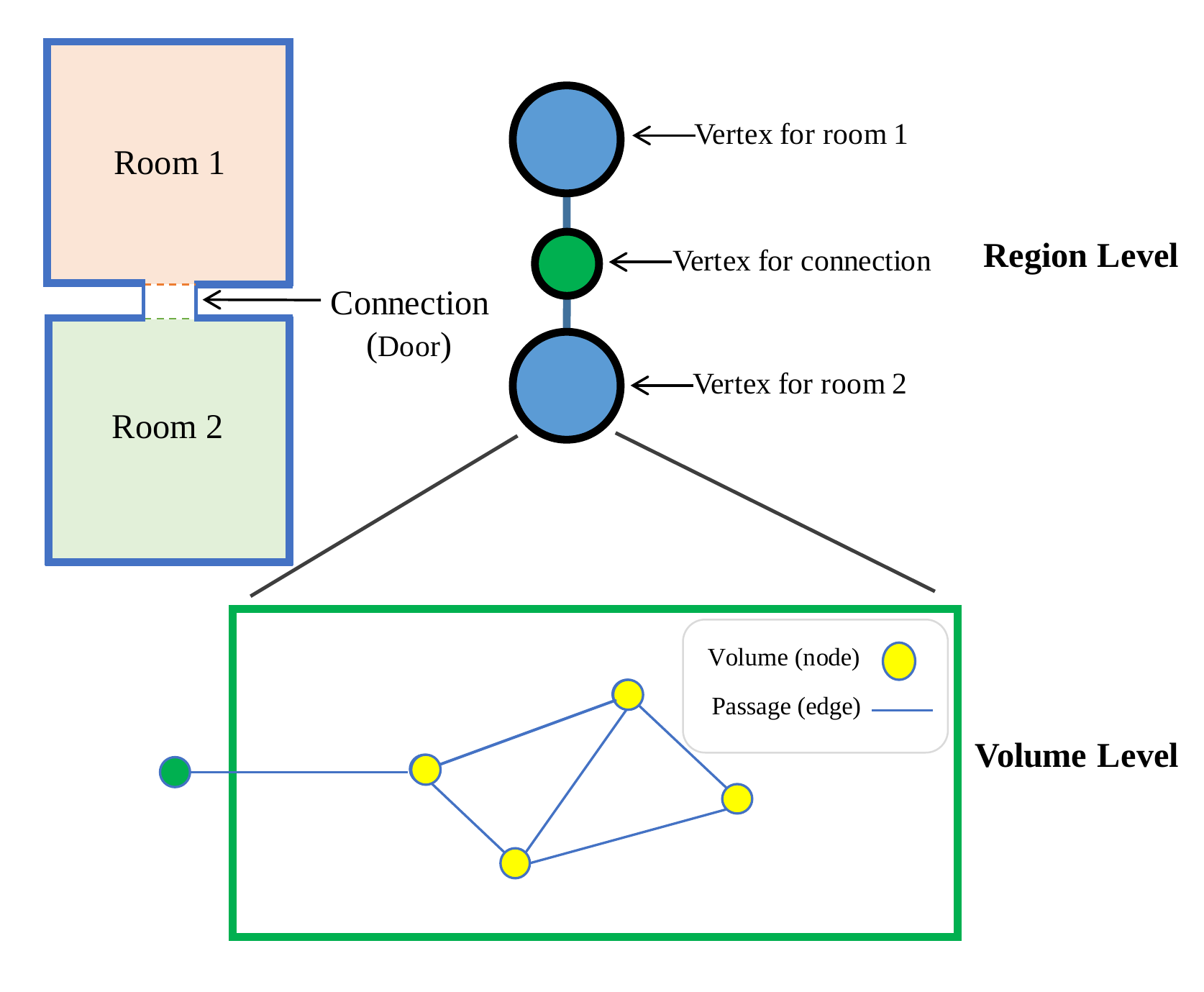}}
	} \\
	\subfloat{
		\fbox{\includegraphics[width=0.9\linewidth]{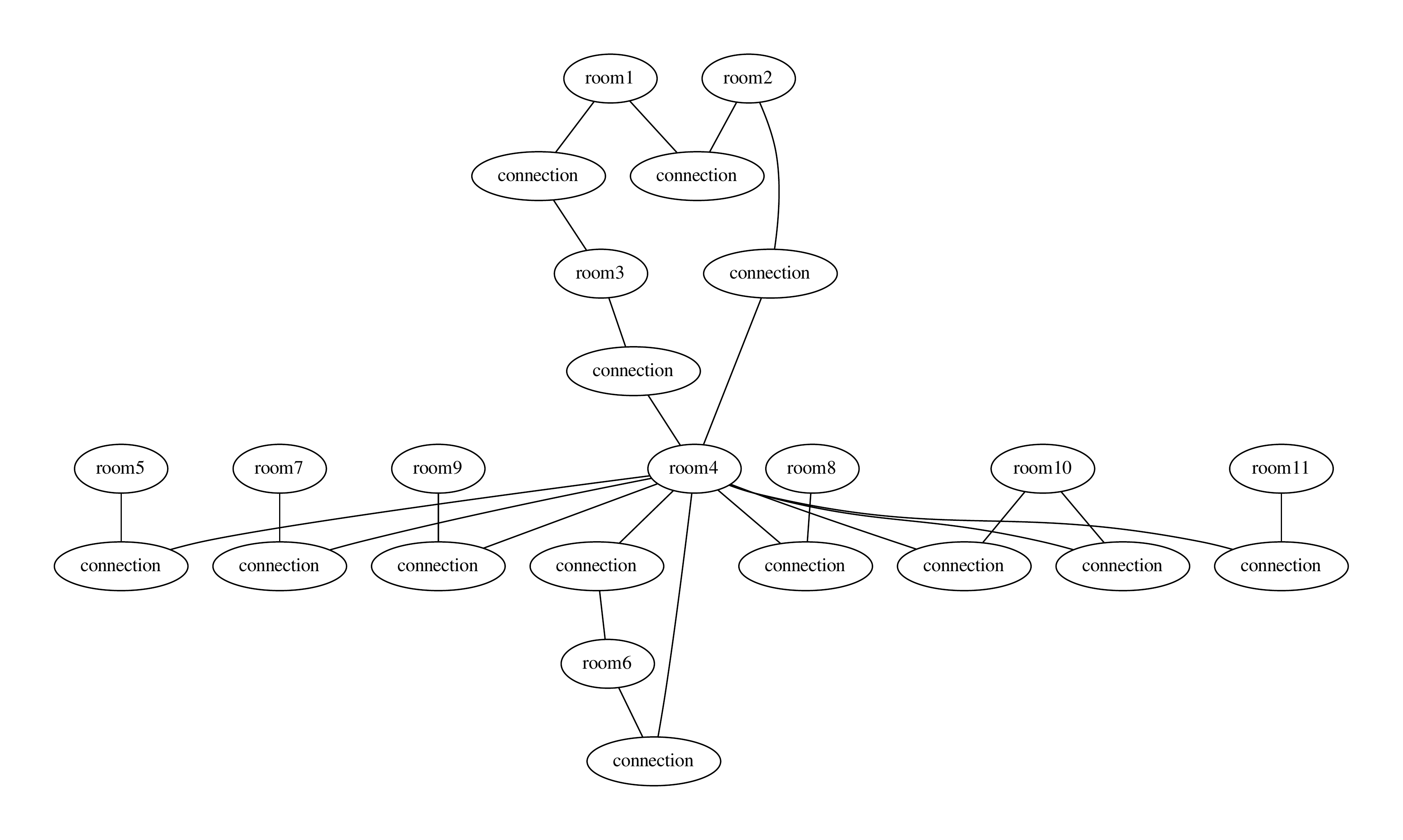}}
	}				
	\caption{\textbf{Top:} The topological graph at $region$ level and volume level. \textbf{Bottom:} One case of the topological graph in the real-world dataset.}
	\label{fig:topo}	
\end{figure}

   \markCM{ 
	An example is shown in Fig. \ref{fig:topo}. There are three vertices at the $region$ level. The two ``room'' vertices correspond to the two rooms, whereas the ``connection'' vertex represents the door. In the next, more detailed, level of the hierarchy, the $volume$ level, the vertices represent the free space clustered together from free space columns of the 3D map. An empty room will typically have just one volume vertex, while tables or other vertical partitions will cause a room to have multiple volume vertices. The door is also a vertex in the region level attributed to ``connection'', representing the traversable volume of the open door. Edges are annotated with $passages$, representing the surface across which one can pass from one $volume$ vertex to the other. Such $passages$ are represented as a mesh. We believe that this mesh representation of the edge/ $passage$ is especially useful for planning and navigation, as it makes available the whole set of points of the passage surface to choose from when planning the passage from one $volume$ to the next.  }

\begin{figure*}[ht]
	\includegraphics[width=\linewidth]{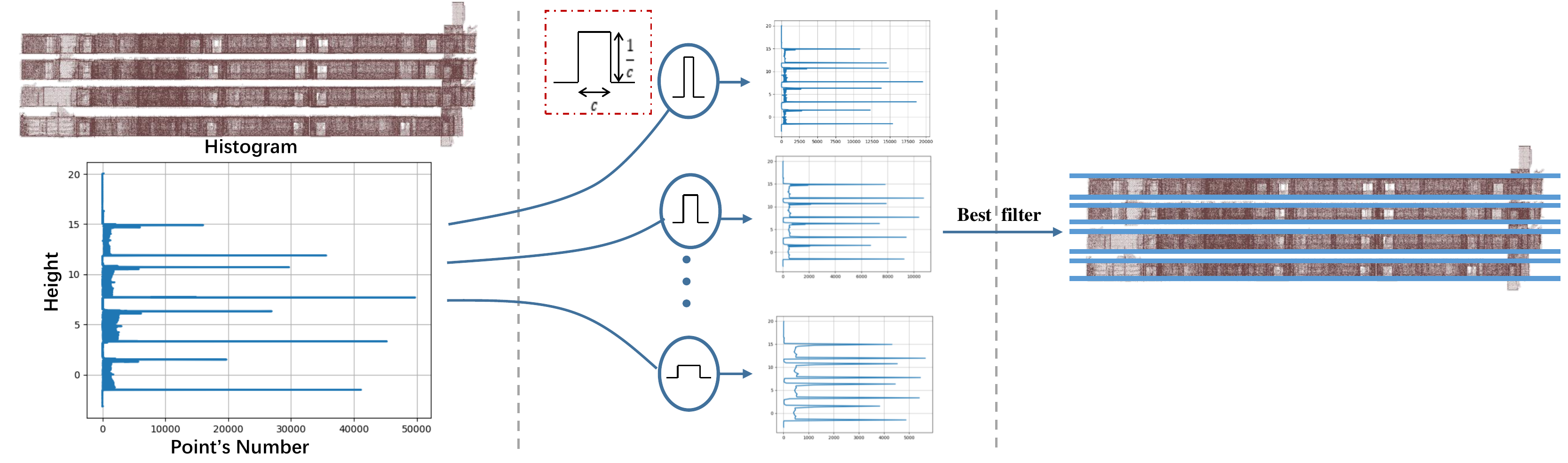}					
	\caption{\textbf{Left:} The point cloud and the histogram of the $z$-$axis$. \textbf{Middle:} Some of the window filters used, from top to the bottom the window size in cm is $c\in\{2,4,6,8,10\}$.  \textbf{Right:} The blue lines show the position of ceilings and floors.}
	\label{fig:storey}
\end{figure*}     
	
\subsection{Input Point Cloud Preprocessing}
	\label{method:storey}
	The input of our algorithm is a complete 3D indoor building point cloud. We require that the ``$up$'' direction of the real-world is the $z$-$axis$ of the point cloud\footnote{\label{f:footnote} If this requirement is not met, the z-axis normal could be estimated by Principal Component Analysis (PCA). There are also other methods suitable for this task, such as the Manhattan frame estimation proposed by \citet{ghanem2015robust} for more complicated cases.}. \markCM{ In the first step we use a voxel filter in the Point Cloud Library(PCL), \citet{rusu20113d} with a voxel size of 5cm to limit the density of the point cloud. }

	\markCM{
	Point clouds often contain noise and erroneous points. A major source of such noise is the reflection of the beams of the LiDAR sensor on glass or other reflective material, \cite{zhao2020reflection}. To denoise the input point cloud, we use a filtering step. We utilize the clustering technique with Euclidean metric from the PCL\footnote{Euclidean Cluster Extraction}. If the distance between points is less than a specific threshold, they will be clustered together. For our experiments we choose a threshold of 20cm. After clustering we reject all clusters with less than 100 points. In Table \ref{table:fliter}  we show the number of $volumes$  for different datasets with and without the filtering process. After rejecting the clusters with fewer than 100 points, we can reduce the number of $volumes$ which stem from noise and erroneous points.}

	\begin{table}[h!]
		\begin{tabular}{@{}
				>{\columncolor[HTML]{FFFFFF}}c 
				>{\columncolor[HTML]{FFFFFF}}c 
				>{\columncolor[HTML]{FFFFFF}}c 
				>{\columncolor[HTML]{FFFFFF}}c @{}}
			\toprule
			\# Volumes & Before filtering & After filtering & Ratio(\%)     \\ \midrule
			Dataset 1       & 7,874             & 7,496            & \textbf{95.2} \\
			Dataset 2       & 1,103             & 1,087            & \textbf{98.5} \\
			Dataset 3       & ~~157              & ~~145             & \textbf{92.4} \\
			Dataset 4       & ~~893              & ~~890             & \textbf{99.7} \\
			Dataset 5       & ~~490              & ~~388             & \textbf{79.2} \\
			Dataset 6       & ~~855              & ~~528             & \textbf{61.8} \\ \bottomrule
		\end{tabular}
		\caption{The number of volumes in different datasets with and without the filtering process.}
		\label{table:fliter}
	\end{table}

    Our method can deal with point clouds from multi-storey buildings. Inspired by \citet{armeni20163d}, we use a peak detector to recognize the height of the floor from the unfiltered point cloud. Since only the surface of objects is visible to a 3D sensor, the space above the ceiling surface and below the floor surface is empty (no points here). If we form a 1-dimensional histogram of the $z$-$axis$, peaks are expected at floor and ceiling height. We use a space filter to ensure that the input point cloud is uniform. The floors and ceilings are then detected in the $peak$ pattern.    
    We assume that the floor in the building is horizontal, because sloping or uneven ground only accounts for a small part of the interior space layout, \cite{steadman2006most}.
    
    We use a set of window filters combined with Otsu's method, \cite{otsu1979threshold} to find the best floor and ceiling candidates without setting the threshold manually. The filters are represented as: $g_c(s) = \frac{1}{c} \Pi_c(s),$ where $\Pi_c(s) = \mathbb{I}[|s| \leq c]$ and $ \mathbb{I}[A] $ are indicator functions with 1 when A is true and 0 otherwise. 
    
    An example density histogram signal is shown in Fig. \ref{fig:storey} (left). There may be multiple peaks at ground height due to noise near the ground and a possible sloping of the outside of the ground around the building. Different window sizes combined with Otsu's method produce different results, as shown in Fig. \ref{fig:storey} (middle). \markCM{ Some of these candidates may not correspond to actual heights of ceilings or floors, so we use a clustering method to find the best window size. Window sizes are clustered together if a) they have the same number of peaks and b) those peaks overlap in the histogram. 	The cluster with the highest number of window sizes is selected, and then the peaks from the smallest of those window sizes are used because a smaller window size has a higher spatial resolution. The center of each peak is then selected as the peak position. Finally, the peaks are labeled as floor and ceiling from the bottom up in an alternating manner. An example segmentation result is shown in Fig. \ref{fig:storey} (right), where blue lines indicate peak positions. }

	\subsection{Column Generation}
	\label{method:column}
	
	The 3D voxel occupancy map divides the space into compact voxels with one binary attribute: occupied and unoccupied. Occupied indicates that there are points in the voxel, and unoccupied shows that the voxel is empty. With this we can transfer the point cloud into a 3D voxel occupancy grid map by using the open-source tool $sdf\_tools$ from the The Autonomous Robotic Manipulation Lab, University of Michigan\footnote{\url{https://github.com/UM-ARM-Lab/sdf_tools}} \citet{hayne2016considering}. 
	
	In the voxel occupancy map, our method finds $columns$ by iterating over the $x$-$axis$ and $y$-$axis$. The $column$ represents free space that has the same length and width as a voxel in the map, while the height is determined by the continuous free space. \markCM{ For a given $x$ and $y$, going from the top to the bottom of the 3D voxel occupancy map, we merge adjoining free voxels into $columns$. \linebreak $Columns$ not starting from or stopping at occupied voxels (i.e. the top-most and bottom-most $column$) are removed to only represent the free space inside the building. } Each $column$ is attributed with the bottom height $z_1$, the top height $z_2$, and the $x, y$ position. A simple case is shown in Fig. \ref{fig:column} (Left).
	
	When a high-precision laser scanner generates a point cloud map, it is usually impossible to see the ground directly under the scanner, so point clouds could have gaps where the floor should be. We assume that the floor in the building is horizontal so that we can use the height of the floor from the previous step to constrain the $column$ generation process. In detail, the generation process, which generates $columns$ from top to bottom, stops when it reaches the height of the floor of the current storey. Since the height of the ceiling does not need to be consistent in the indoor environment, a $column$ with infinite height is treated as an exterior $column$ and deleted.
	
	\subsection{Volume and Passage Generation}
	\label{method:volume}  
	
\begin{figure}[t!]
		\subfloat[ \textbf{Left}:  All $volume$ vertices and their connectivity in topological graph (D1), $volume$s with size more than $20m^3$ are shown in blue. \textbf{Right}: All the $volumes$ in 3D voxel occupancy map (D3).]{
			\label{fig:volume}
			\fbox{\includegraphics[width=0.46\linewidth]{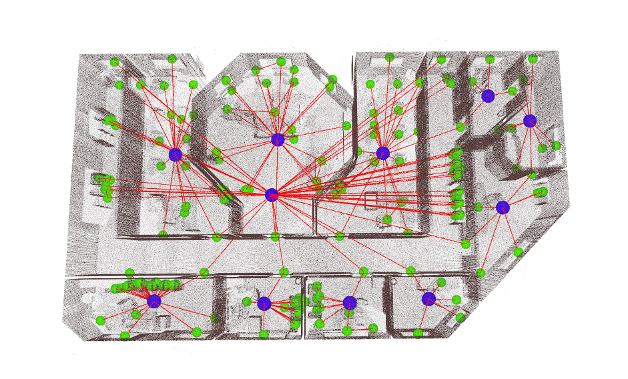}
				\includegraphics[width=0.46\linewidth]{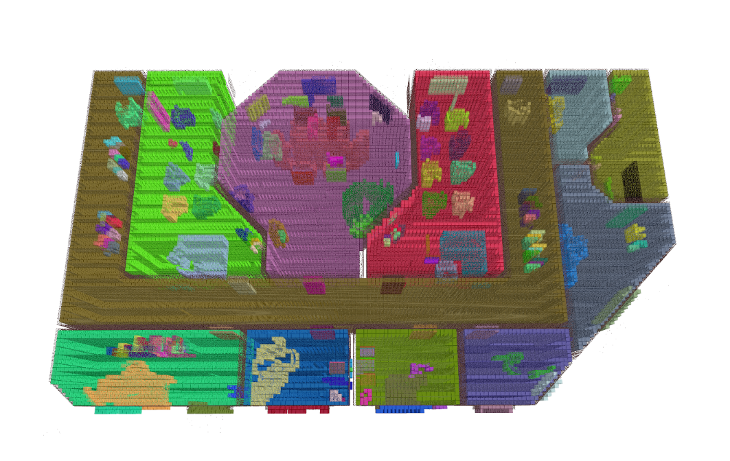}}
		} \\
		\subfloat[ \textbf{Left}: The $region$ vertices (D1) in blue and $passage$s in green. \textbf{Right}: The $regions$ in 3D voxel occupancy map (D3).]{
			\label{fig:region}
			\fbox{\includegraphics[width=0.46\linewidth]{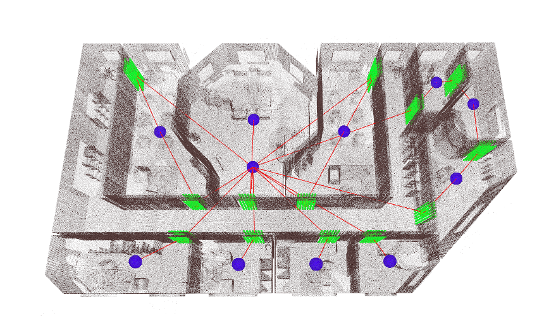}
				\includegraphics[width=0.46\linewidth]{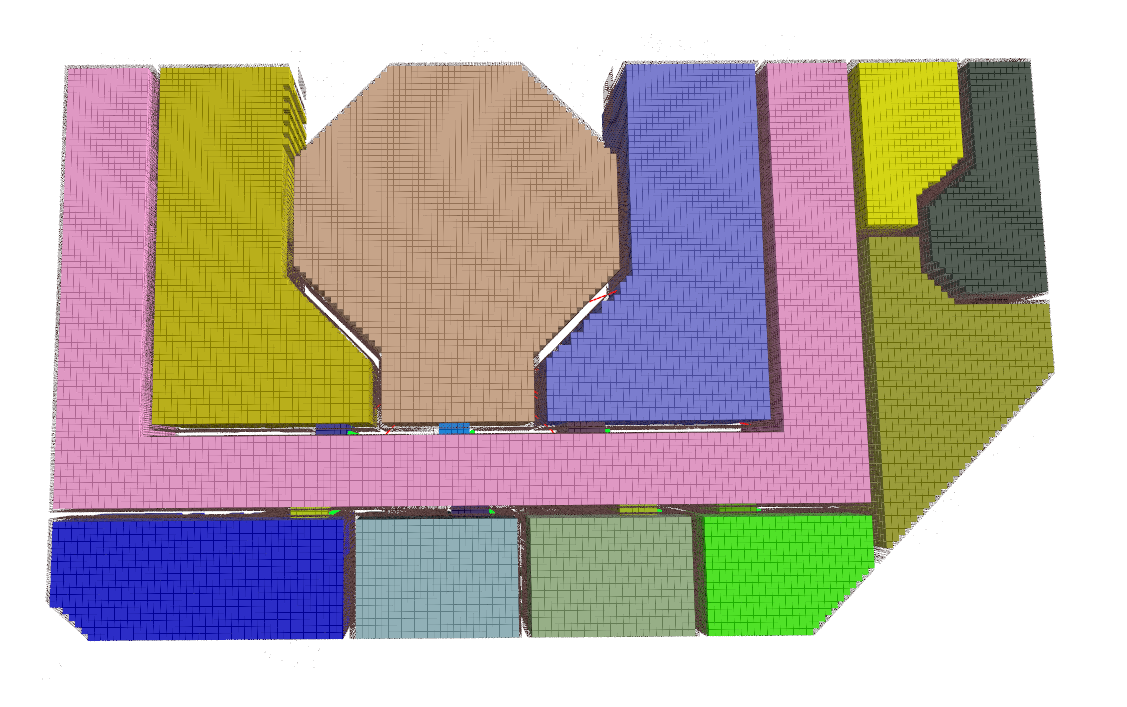}}
		}							
		\caption{Hierarchical topological representation. }
	\end{figure}

	\markCM{$Volumes$ are assembled from contiguous $columns$. } A schematic diagram is shown in Fig. \ref{fig:column} (middle) and a real case is shown in Fig. \ref{fig:volume} (right) with different colors for different $volumes$. $Volumes$ are generated as follows: We randomly pick a $column$ as a seed and perform an iterative search for neighboring $columns$ that are similar to it. The similar $columns$ need to have a similar top height (we chose $10\%$ of the length of $columns$ as difference threshold). With this, our $column$ merge process is suitable for surfaces of continuously varying height (e.g. slanted ceilings that are not too steep). Spaces with large local variations in height are divided into smaller $volumes$.
	
	There is no requirement for the bottom height of the merged $columns$ to be continuous. This is because we aim to generate big $volumes$ for rooms, that pass over obstacles like furniture. Those big $volumes$ are then excellent seeds for the $region$ generation, which should encompass the whole room. The potential dis-continuous nature of the bottom of some $volumes$ means, that they cannot be immediately used for ground robot navigation. This is because the emphasis of this work is on 3D volumetric representation. But an easy extension of our method could add another hierarchy level at the $volume$ level, where $volumes$ are split to ensure that all volumes are continuous at the bottom. $Passages$ between $volumes$ could then be annotated with information if they are continuous (i.e. the connection of the two $volumes$ starts at the same height).

\begin{algorithm}[t]
	\caption{Passage generation algorithm}
	\label{alg:PG}  
	
	\KwIn{ Set of $column$ vertices attributed with their $volume$ id: $CV$ }
	
	\KwOut{ Set of edges for the $volume$ graph with attributed $passage$s: $\mathcal{E}$ }
	
	\Begin{
		
		$P = \{P^{i}_{j} = \emptyset |\ i, j \in volume\_indices \}$ \noteCM{ // $P$ is the set of all possible $passage$s, which are all empty sets at the beginning}
		
		\For{each $cv \in CV$} {
			
			\For{each $n \in neig(cv, CV)$}{ \noteCM{ // go through all neighbors of $cv$ } 
				
			\If{$vid(cv) \neq vid(n)$}{ \noteCM{ // $vid(\cdot)$ returns the $volume$ id of a $column$. If the $volume$ ids of $cv$ and $n$ are different... }
		       
		       $g = cell\_pts(cv) \cap cell\_pts(n)$ \noteCM{ // $cell\_pts(\cdot)$ returns the set of grid points of all cells of a $column$ (8 grid points per cell); This line returns all points shared between both columns: the $passage$ points }
		       
		       $P^{{vid(cv)}}_{{vid(n)}} = P^{{vid(cv)}}_{{vid(n)}} \cup g$ \noteCM{ // Add the found $passage$ points to the according set }
		    }
		}
	}

	\For{each $P^i_j \in P$}{
		\If{$P^i_j \neq \emptyset$}{
			$SC = clustering(P^i_j, d_{th})$ \noteCM{ // clustering algorithm returns a set of clusters = $passage$s }
			
			\For{each $sc \in SC$}{
				$mesh(sc) \rightarrow e^i_j$ \noteCM{ // generate a mesh from $sc$ and attribute it to the edge $e^i_j$ }
				
				$\mathcal{E} = \mathcal{E} \cup \{ e^i_j \}$ \noteCM{ // add this edge/ $passage$ to the graph }
			}
		}
	}
}
\end{algorithm}

\begin{algorithm}[t]
	\caption{clustering}
	\label{alg:CL}  

	\KwIn{ Set of contact surface points between $V_i$ and $V_l$: $S^{V_i}_{V_l}$ 
			
			Distance threshold: $d_{th}$
		}
		
	\KwOut{ Set of clusters: $SC$ }
	
	\Begin{
		
		$Q = S^{V_i}_{V_l}$  
		
		\While{$Q$ not empty}{
			$p_1 \in Q$ \noteCM{ // select a point $p_1$ from $Q$ }
			
			$Q = Q \setminus \{p_1\}$ \noteCM{ // remove $p_1$ from $Q$ }
			
			$F = F \cup \{p_1\}$ \noteCM{ // add $p_1$ to frontier $F$ }
			
			$C = C \cup \{p_1\}$ \noteCM{ // ad $p_1$ to cluster $C$ }
			
			\While{$F$ not empty}{
				 $p_2 \in F$ \noteCM{ // select a point $p_2$ from $F$ }

				$F = F \setminus \{p_2\}$ \noteCM{ // remove $p_2$ from $F$ }
				
				$N = \{p \in Q | $ $ geom\_dist(p, p_2) < d_{th} \}$ \noteCM{ // $ geom\_dist$ returns the geometric distance between two points. This line creates a set $N$ of all points in $Q$ close to $p_2$. }
				
				$Q = Q \setminus N$ \noteCM{ // remove $N$ from $Q$ }

				$C = C \cup N$	\noteCM{ // add $N$ to $C$	}	
				
				$F = F \cup N$ \noteCM{ // add $N$ to $F$ }
			}
			$SC = SC \cup \{C\}$ \noteCM{ // add $C$ to set of clusters $SC$ }
        }
		return $SC$
	}
\end{algorithm}
    
    An edge represents a traversable $passage$ connecting two $volumes$. Each $passage$ is generated from the contact surface shared by two $volumes$. 
    \markCM{ Considering our topological map in $volume$ level is a graph $\mathcal{G}\triangleq(\mathcal{V},\mathcal{E})$ (where the vertices $\mathcal{V}$ are the $volume$ and the edge set $\mathcal{E}$ includes the $passage$ between $volume$s). See Alg. \ref{alg:PG} for the pseudocode of generating $passage$ and edge in the graph  $\mathcal{G}$. In the first \textbf{For} loop, we extract the 3D grid points of the contact surface between different $volume$ and put them into the possible $passages$ set $P$. There could be different $passages$ between two $volumes$, so we use a clustering algorithm (see Alg \ref{alg:CL}) to find sets of $passage$ points from each element of $P$. A mesh is generated from each of these sets of $passage$ points and attributed to the corresponding edge. This edge is then added to the graph  $\mathcal{G}$. }  In  Fig. \ref{fig:region} (left) we show the $passages$ at the $region$ level in green.

\subsection{Region Generation from Volumes}
	\label{method:region}
	
\begin{algorithm}[tb]
	\caption{Region Generation}
	\label{alg:RG}  
	
	\KwIn{ $\mathcal{G}$, $a_{th}$}
	
	\KwOut{ set of regions $R$ : $RS$ }
	
	\Begin{
		
		\For{each $ v \in \mathcal{V} $} {  \noteCM{ // go through each $volume$ $v$}
			
			$Count(v)$ = 0, $S$ = $\{\}$, $RS$ = $\{\}$ \noteCM{// initialize seeds set $S$ and regions set $RS$, $Count(\cdot)$ store how many $R$ contain the input $volume$.}
			
			\If{$size(v) > a_{th}$ } { \noteCM{ // $size(\cdot))$ return the size of input $volume$ }
				
				$S \cup \{v\}$  \noteCM{// add the $v$ to $S$}
			}
		}

		$S\_filter = filtering(S)$  \noteCM{// $filtering(\cdot)$ returns clusters of $S$, two seeds that have edge between them will be clustered together. This line merges those seeds that have edge between them into one set.}

		\For{each $sc \in S\_filter$}{  \noteCM{// go through each seed cluster}
			
			$R$ = $\{\}$, $F$ = $\{\}$
			
			$R = R \cup sc$ \noteCM{ // add $sc$ to region $R$}
			
			$F = F \cup sc$ \noteCM{ // add $sc$ to frontier $F$}

			\While{$F$ not empty}{
				
				$v_1 \in F$ \noteCM{//select a $volume$ $v_1$ from $F$}
				
				$F \setminus \{v_1\}$ \noteCM{// remove $v_1$ from F}
				
				$N = \{ v \in \mathcal{V} |\ edge\_between(v_1, v) = true, \ v \notin S,\ v \notin R \}$  \noteCM{ // $edge\_between(\cdot,\cdot)$ return true if there is a edge in $\mathcal{G}$ between two $volumes$. This line create a set $N$ store all the $volume$s connected with $v_1$ while not in $S$ and $R$. } 
				
				$R = R \cup N$ \noteCM{// add $N$ to $R$}
				
				$F = F \cup N$ \noteCM{// add $N$ to $F$}
			}
		
			\For{each $v$ in $R$}{
				
				$Count(v)$ += 1 \noteCM{// This line counts the number of $R$ that contain $volume$ $v$.}
			}
		
			$RS = RS \cup \{R\}$ \noteCM{// add $R$ to $RS$}
		}
	
		\For{each $ v \in \mathcal{V} $} {
			
			\If{$Count(v) > 1$} {   	
				
				remove $volume$ $v$ from all set $R$ in $RS$. 
				
				$RS = RS \cup \{\{v\}\}$ \noteCM{// add $\{v\}$ to $RS$}
			}
		}		
		
	}
\end{algorithm}

	The next step is generating the graph at the $region$ level (Fig. \ref{fig:column} (right)), which represents enclosed spaces such as rooms. \markCM{ Vertices in the $region$ level have one binary attribute, ``room'' or ``connection''. Typically, ``room'' indicates the room or the corridor and ``connection'' indicates the door.  Fig. \ref{fig:region}(right) shows different $region$s in different colors - the regions below the doors are there, but occluded. }

	\markCM{ In indoor environments, we note that rooms connect with other places by doors. We also note that the room's size is much bigger than door's size. So the goal of $region$ generation method is to find the ``room'' - ``connection'' - ``room'' pattern in graph $\mathcal{G}(\mathcal{V},\mathcal{E})$. We chose those $volumes$ that exceed the size threshold $a_{th}$ as ``room'' seeds and generate a $region$ form each seed.  The connection part between two $regions$ is then extracted as a single $region$, attributed with ``connection''.   }
	
\begin{figure}[h!]
		\centering
		\fbox{\includegraphics[width=0.95\linewidth]{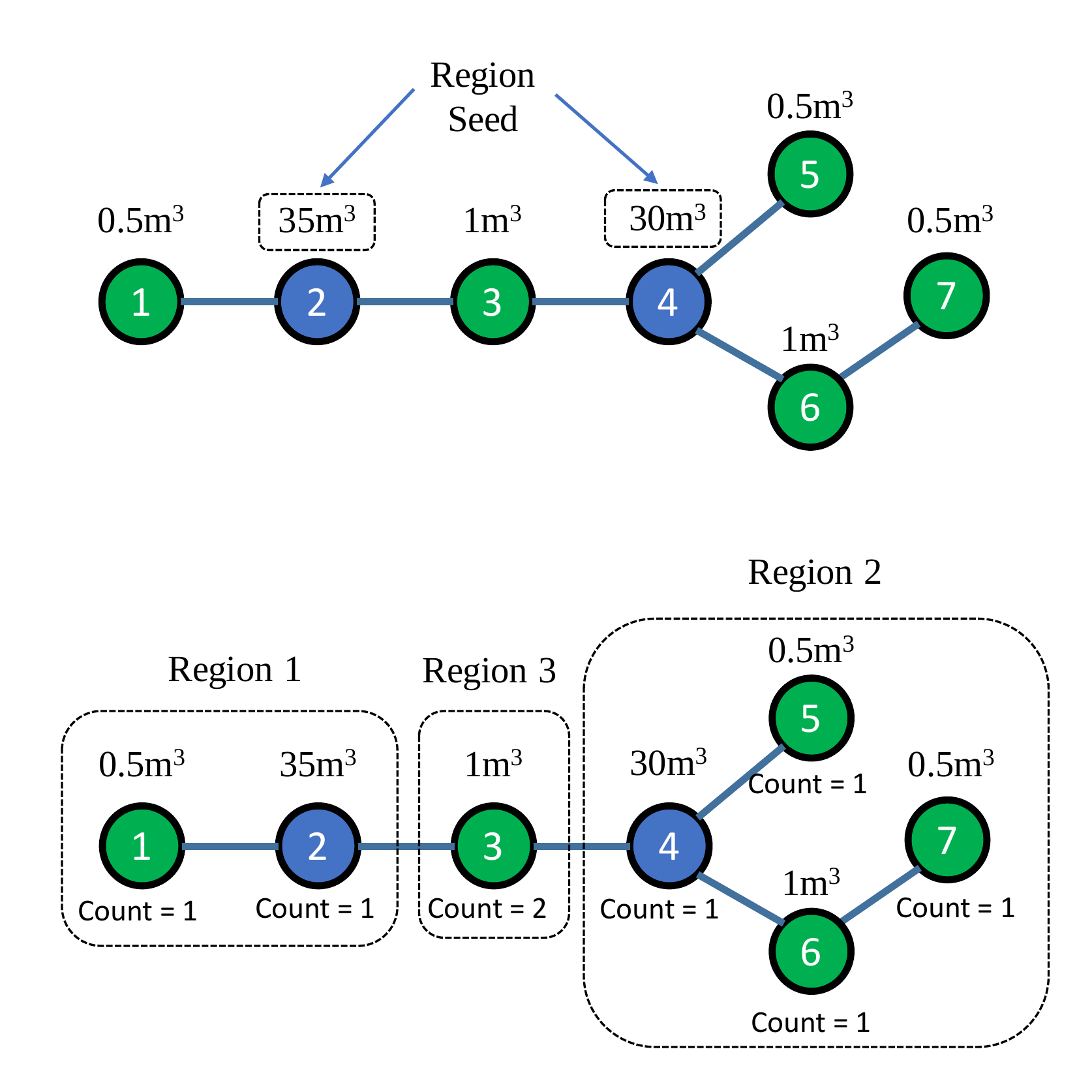}
		}						
		\caption{Region generation from $volumes$.}
		\label{fig:region_generation}
	\end{figure}

	\markCM{ Alg. \ref{alg:RG} shows the detail of our method. To boost our performance, we use a filtering algorithm (line 8) to remove extra seeds in one room. Any two seeds with an edge between them will be merged together. We use a breadth-first search that starts from each seed cluster $sc$ and stops when it meets other seeds to generate $regions$. But some of the $regions$ can have connection volumes, such as the Vertex 3 in Fig. \ref{fig:region_generation}, which belong to two $regions$ (simultaneously generated from Region Seed 2 and Region Seed 4). In $Count$ we store the number of regions that each $volume$ belongs to. The $volumes$ with a $Count$ bigger than 1 are removed from all $regions$ and are subsequently added as a new individual $regions$, attributed with ``connection'' (Region 3 in Fig. \ref{fig:region_generation}). }
	
\begin{figure}[h!]
	\subfloat{
		\fbox{\includegraphics[width=0.90\linewidth]{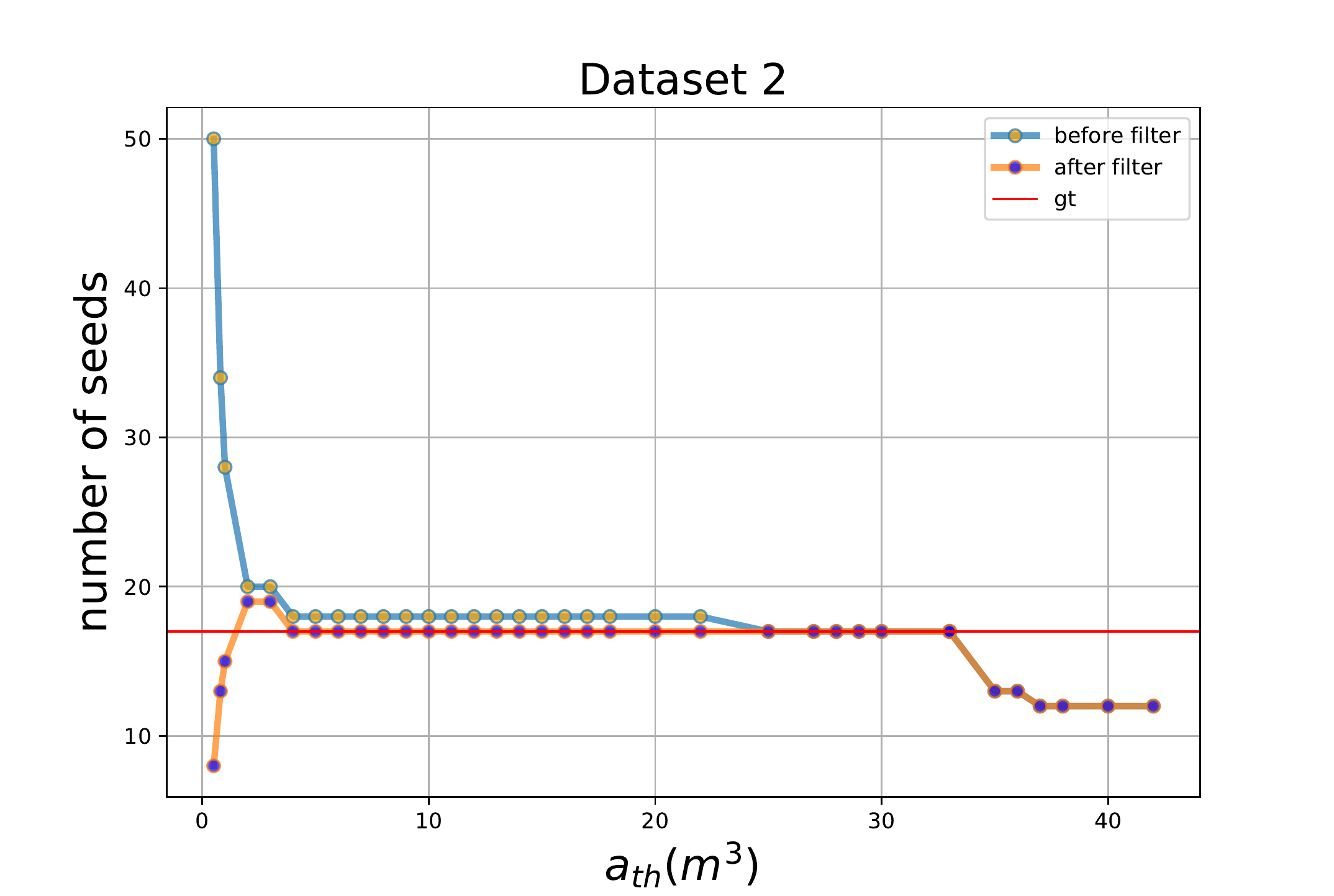}
		}
	}							
	\caption{Threshold evaluation. Origin dot show seeds' number without filtering step. Blue dot show seeds' number after filtering. Red line show the ground truth regions' number. The numbers of seeds should equal with number of gt.}
	\label{fig:threshold_eva}
    \end{figure}

	\markCM{ Here we discuss the size threshold $a_{th}$ and the seeds filter step in the pseudocode. An obvious selection rule of threshold $a_{th}$ is for it to be larger than the size of the door and smaller than the size of smallest room. Typically, a $volume$ generated from a door will not exceed $1 m^3$. A small $a_{th}$ can lead to many seeds in one room, so we use a filter step to merge those seeds which are connected with each others. The idea behind this step is that regions (e.g. room, corridor) only connect with other regions by doors. Hence the door's $volume$ should not be selected as a seed, and the filter step should only merge seeds in the same region rather than combine seeds from different regions. This step makes the selection of $a_{th}$ more flexible and robust, as shown in Fig. \ref{fig:threshold_eva}. In this small experiment we test different values for $a_{th}$ in Dataset 2. In the beginning, $a_{th}$ is smaller than the size of doors, and we get many seeds. But, as $a_{th}$ increases, the number of seeds decreases quickly. When the $a_{th}$ is bigger than the size of a room, the number of seeds decreases further. We see that the result after the filter step is more stable than the original result and that it is easy to select a suitable $a_{th}$, as a wide range of values (bigger than door, smaller than room) give good results.}
	
	
	\begin{figure}[ht]
		\subfloat{
			\fbox{\includegraphics[width=\linewidth]{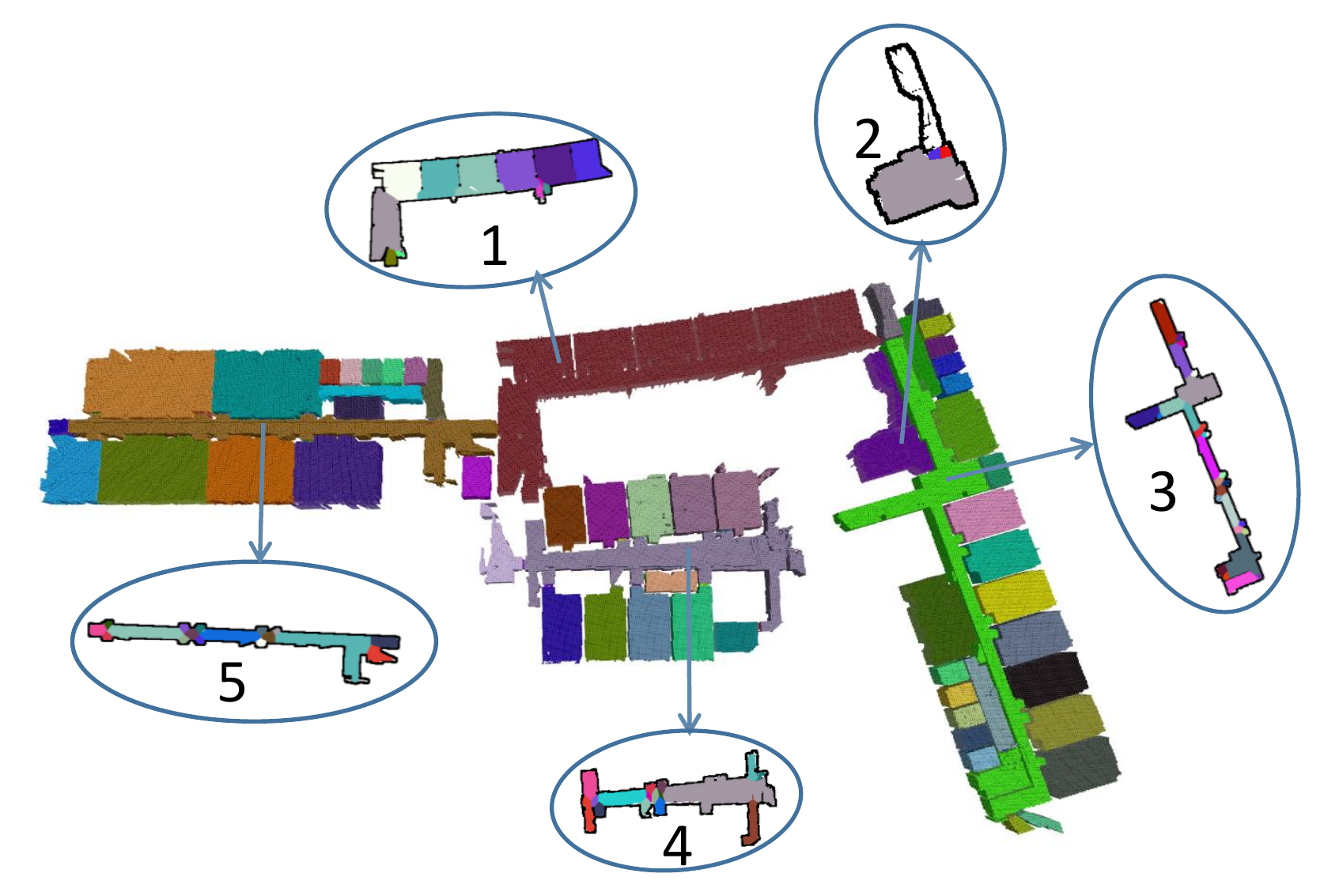}} 
		} \\
		\subfloat{
			\fbox{\includegraphics[width=\linewidth]{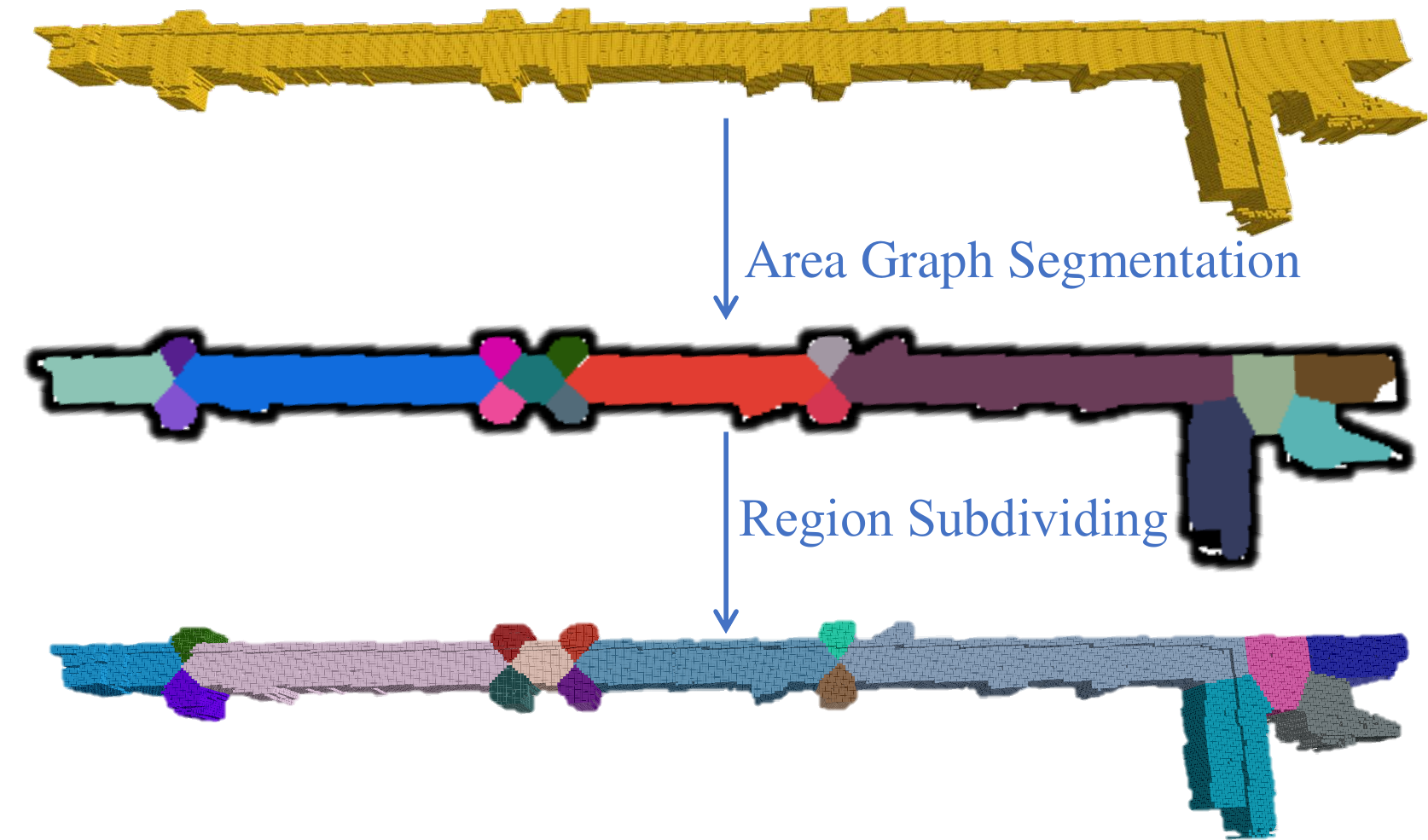}}
		}															
		\caption{\textbf{Top:} For each larger $region$, a 2D picture is generated and segmented using the area graph. \textbf{Bottom:} The 2D result of area graph segmentation is used to segment the $region$ in 3D}
		\label{fig:area}
	\end{figure}

\subsection{Area Graph Segmentation in Region}
	\label{method:area}
    \markCM{ In the previous step, we merged $column$ to $volume$ by height and then $volume$ to $region$ by topological relationship. The results show the expected building structure with rooms and corridors segmented as individual $regions$. However, there are still problems, which we will solve by further sub-dividing regions in another hierarchy level - a more detailed region level. The problems stem from two issues: a) rooms may not be properly segmented into individual rooms due to issues with the sensor or the building structure and b) there may be corridors which are correctly segmented into one region, but for which a further sub-division may be beneficial for some applications. }
    
    \markCM{ In the top of Fig. \ref{fig:area} a big region marked with 1 is shown, which consist of several rooms with ceiling-height glass fronts. The laser scanner did not see this glass, so all the columns of the rooms and the corridor are merged together to one big region, leading to undersegmentation. A similar issue may occur in cases where a door has the same height of the ceiling. }  
    
    \markCM{  Again looking at Fig. \ref{fig:area}, we see corridors marked with 3, 4 and 5, which form three big regions. This may be desired in some cases, but sometimes a finer sub-division may be beneficial: a more fine-grained representation is better to assess the navigation cost during planning. The generated plan is also more detailed, thus making local planning and navigation easier and more reliable. Furthermore, we will end regions at junctions, which are natural decision points of ``which direction to turn to next''. This may be beneficial to human robot interaction, e.g. when specifying a robot path in natural language. }
    
    \markCM{ To tackle these problems, we use a 2D topological segmentation method \citet{hou2019area} to segment big $regions$ in the horizontal direction, called Area Graph. We chose this method over other 2D segmentation methods, because in the paper it showed superior segmentation results when compared to ground truth, while only relying on one one easily selected parameter. We segment $regions$ that exceed a certain $volume$ (e.g. $20 m^3$ ) to 2D grid maps. Then the Area Graph algorithm is applied to the 2D grid map. Since the 2D grid map is generated from the region level, it does not contain any furniture and is thus very suitable for the Area Graph algorithm, which performs best with furniture-free maps, \citet{hou2019area}. }
    
    \markCM{  The cells of the 2D grid map have the same width and length as the $columns$. The $columns$ of the $region$ to be segmented are then projected onto the 2D grid map: Starting with an all ``occupied'' 2D grid map, all coordinates which have a $column$ are marked as ``free''. }
    
    \markCM{  The Area Graph, described in detail in  \citet{hou2019area}, is then applied to the 2D grid map. It is a topological representation where vertices represent areas and edges represent $passages$. It segments the 2D grid map into areas based on the underlying topological structure (i.e. junctions in corridors) as well as detected rooms. There are four main steps to obtain an Area Graph: 
   	\begin{enumerate}
   		\item \textit{Generate a Voronoi Diagram (VD) from the 2D grid map.} The VD contains a set of waypoints with the same distance to the nearest two sites (``occupied'' cells in 2d grid map).
   		\item \textit{Generate Topology Graph, \citet{schwertfeger2015Map}, from VD.} This graph is a topometric representation of the map with vertices in dead-ends of corridors, in junctions and in rooms, and edges connecting them. It is generated by filtering and pruning the VD.
   		\item \textit{Generate Area Graph from Topology Graph.} Edges in the Topology Graph are attributed with an ordered list of waypoints from the VD. Each waypoint has two sites (i.e. ``occupied'' cells with equal and minimum distance to the waypoint).	Using the ``occupied'' cells of a Topology Graph edge and its two end points, a polygon is formed enclosing the free space belonging to the edge - its ``Area'', a vertex in the Area Graph. Passages are generated where polygons (of Topology Graph edges connected to the same vertex) are very close together. The passages are then the edges of the Area Graph.
   		\item \textit{$\alpha$-shape room detection.} Bigger rooms usually have several Topology Graph vertices and edges and are thus oversegmented in the Area Graph. To compensate for this, the $\alpha$-shape algorithm, \citet{edelsbrunner1983shape}, is used to generate polygons in all areas that meet the free space requirements, specified with the parameter called $\alpha$ (see more details below). So every big room will get one individual $\alpha$-shape polygon for its free space. All areas of the Area Graph intersecting with one such $\alpha$-shape polygon are then merged into one big area vertex of the Area Graph, representing one room, thus resolving the aforementioned oversegmentation issue.
    \end{enumerate} }
    
    \markCM{ The $\alpha$-shape is defined on a point set - all occupied cells of the 2D grid map in our case. Two points from this set, together with the diameter $d = \alpha$, define two circles/ disks (for convenience we define the $\alpha$-value as diameter of that circle here). An edge of the $\alpha$-shape is created between these two points only if at least one of the two circles does not contain any other point from the set. All pairs of two points from the set are tested and the created edges are connected to the polygons of the $\alpha$-shape. As discussed in detail in  \citet{hou2019area}, the selection of the $\alpha$ value is important to get the desired segmentation result. 
   	If $\alpha$ is bigger than the width of the door, the circle will not fit through the door, so the rooms and corridors will get individual $\alpha$-shape polygons and thus be properly segmented. As experiments in the paper show, it is thus easy to find a proper value for $\alpha$ following these guidelines:  
   	\begin{enumerate}
   		\item $\alpha$ must be bigger than the (maximum) width of the corridor.
   		\item $\alpha$ need to be smaller than the (minimum) width of the room without conflict with the first rule.
   		\item $\alpha$ ought to be small enough without conflict with the first and second rule.
   	\end{enumerate}
   	The 3\textsuperscript{rd} rule is there to get an $\alpha$ shape as tight around the room as possible. But in fact the Area Graph is not very sensitive to this, since the actual area polygon is from the merged area vertices, not the $\alpha$-shape polygon. As said above, the $\alpha$ polygon is just selecting which Area Graph nodes to merge, based on whether the polygons of the $\alpha$-shape and the Area node overlap. }
    
    As shown in Fig. \ref{fig:area} (top), for the large $region$, a 2D picture is generated and segmented using the Area Graph algorithm. The segmentation result will be reapplied to the $region$ for a subdividing (Fig. \ref{fig:area} (bottom)). \markCM{ This process is shown in Alg \ref{alg:RS}. We split the $volumes$ and put them into a new $region$ according to the color classifications. Then the big $region$ is the parent of multiple small $regions$ in the $region2$ hierarchy level. }  Afterwards the edges between these small $regions$ and the edges to other $regions$ will be generated by the passage generation algorithm presented in Alg \ref{alg:PG}.

    \begin{algorithm}[t]
    	\caption{Region Subdividing}
    	\label{alg:RS}  
    	
    	\KwIn{ $region$ : $R$, correspondence image: $I$ }
    	
    	\KwOut{ set of $regions$: $SR$ }
    	
    	\Begin{
    		
    		$SR$ = $\{\}$  \noteCM{ // initialize  $sub$-$regions$}
    		
    		\For{each $color \in I$} {  \noteCM{ // go through each $color$ }
    			
    			$S(color)$ = $\{\}$   \noteCM{ // $S(\cdot)$ return link to $region$ with this input color. Initialize the $region$ with no $volumes$ here. }
    			
    		}
    		
    		\For{each $volume$ $V$ $\in$ $R$} {
    			\noteCM{ // $volume$ $V$ is a set of $columns$}
    			
    			$Q$ = $V$
    			
    			\While{$Q$ is not empty} {
    				
    				$column$ $c_1 \in Q$ \noteCM{ // select a $column$ $c_1$ from $Q$ }
    				
    				$color_1$ = $get\_color(I, c)$ \noteCM{ // $get\_color(I, c_1)$ get the color from image $I$ according to the 2D coordinate (x,y) of $c_1$ }
    				
    				$N = \{ c \in Q |\ get\_color(I, c) == color_1 \}$ \noteCM{ find all $columns$ in $Q$ with the same $color_1$ and save as new $volume$ N}
    				
    				$S(color)$ = $S(color)$ $\cup$ $\{N\}$  \noteCM{// add $N$ to $region$ $S$ }
    				
    				$Q$ = $Q$ $\setminus$ $N$ \noteCM{ // remove $N$ from $Q$ }
    				
    			}
    			
    		}
    		
    		\For{each $color$ in $I$} {
    			$SR$ = $SR \cup S(color)$  \noteCM{ // add $S(color)$ to $SR$ }
    		}
    		
    		return $SR$
    	}
    	
    \end{algorithm}
    
    \begin{table*}[h!]
    	\centering
    	\begin{tabular}{ | c | m{4.5cm} | m{4.4cm} | m{4.2cm} | }
    		\hline
    		& \makecell{Dataset 1} & \makecell{Dataset 2} & \makecell{Dataset 3}\\ \hline

    		\makecell{ \textbf{Input} \\ points \\ boundary (x/y/z) \\ regions }
    		& \makecell{ 12,956,732 \\ 169.80m/104.41m/5.01m \\ 70 }
    		& \makecell{ 2,929,680 \\ 73.57m/28.97m/9.33m \\ 18} 
    		& \makecell{ 5,099,751 \\ 11.25m/6.98m/3.23m \\ 7} \\ \hline
    		
    		\makecell{ \textbf{Info} \\ Voxel size \\ Time (seconds) \\ $a_{th}$ ($m^3$) } 
    		& \makecell{ 0.15 \\ 3.10 \\ 20 } 
    		& \makecell{ 0.15 \\ 0.73 \\ 20 } 
    		& \makecell{ 0.15 \\ 0.058 \\ 2 } \\ \hline
    		
    		\makecell[c]{  \\ \\ \\ \textbf{Input} \\ \\ \\ \\ \\ \\ \\  \textbf{3D result}  \\ \\ \\ \\ \\ \\ \\ \textbf{1D result} }
    		& 
    		\begin{minipage}{.23\textwidth}
    			\subfloat{\includegraphics[width=1.0\linewidth, trim=0 0 0 350]{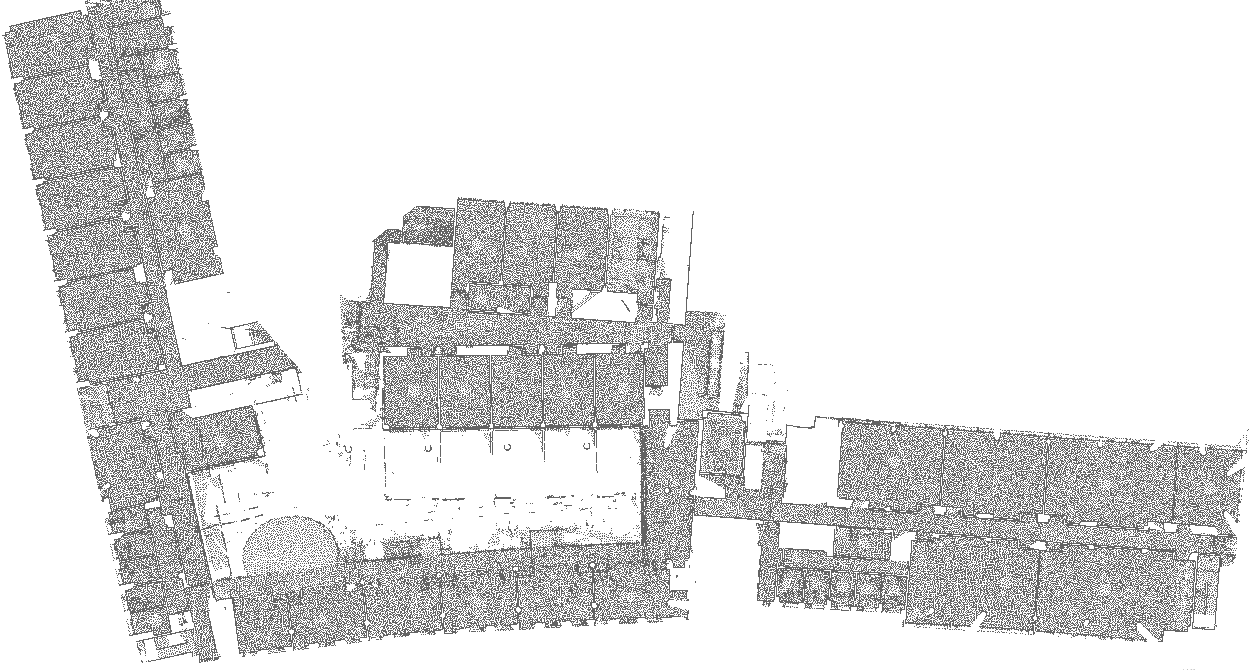}} \vfill
    			\subfloat{\includegraphics[width=1.0\linewidth]{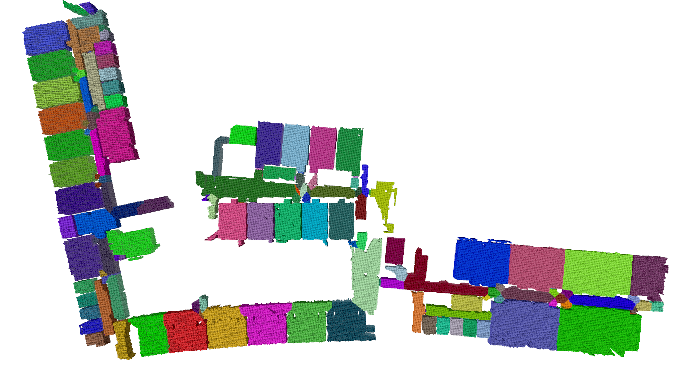}} \vfill	
    			\subfloat{\includegraphics[width=1.0\linewidth]{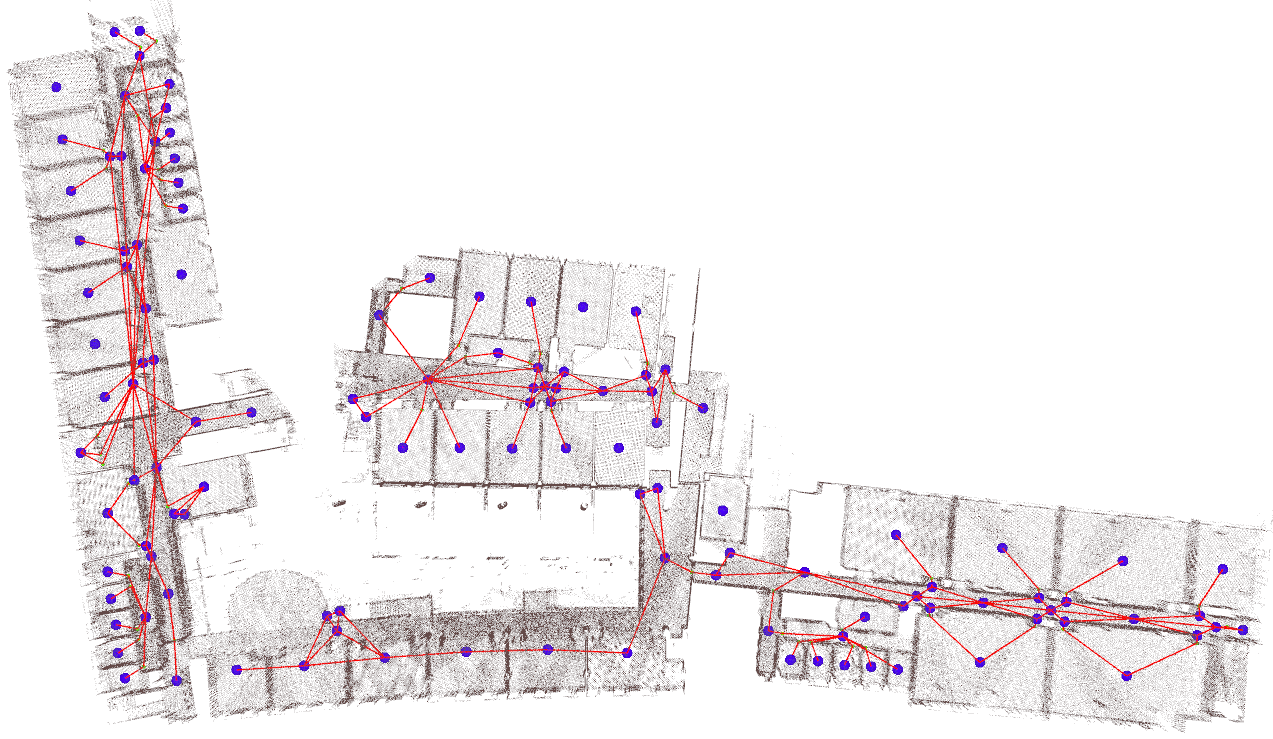}}	
    		\end{minipage}
    		& 
    		\begin{minipage}{.23\textwidth}
    			\centering
    			\subfloat{\includegraphics[width=0.90\linewidth, trim=0 0 0 150]{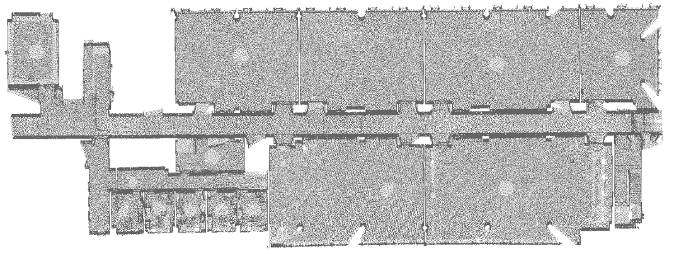}} \vfill	
    			\subfloat{\includegraphics[width=0.90\linewidth, trim=0 0 0 -150]{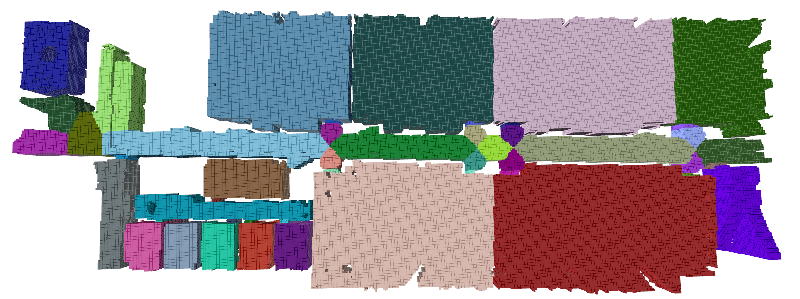}} \vfill
    			\subfloat{\includegraphics[width=0.90\linewidth, trim=0 0 0 -150]{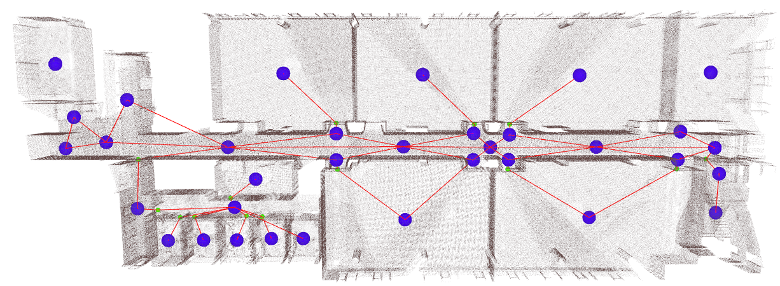}}
    		\end{minipage}
    		&
    		\begin{minipage}{.23\textwidth}
    			\centering
    			\subfloat{\includegraphics[width=0.70\linewidth, trim=0 0 0 150]{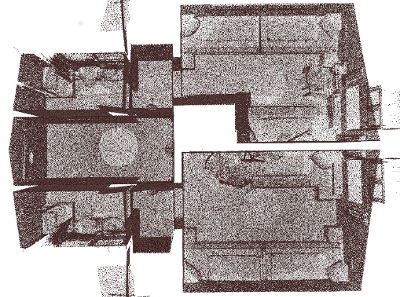}} \vfill	
    			\subfloat{\includegraphics[width=0.70\linewidth]{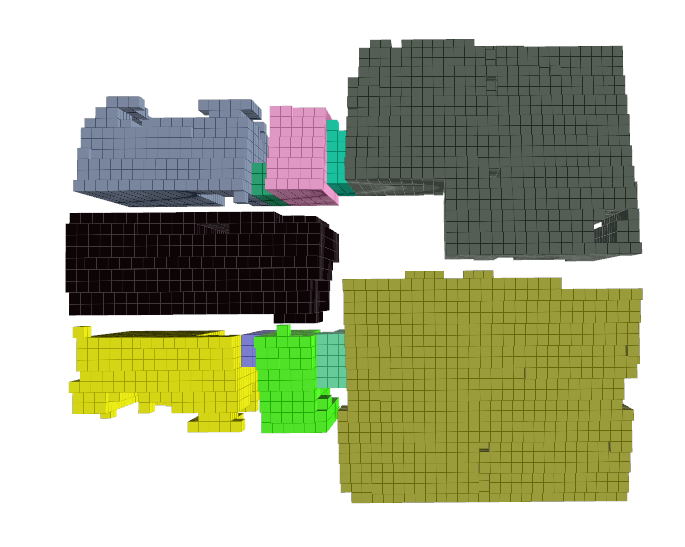}} \vfill
    			\subfloat{\includegraphics[width=0.70\linewidth, trim=0 20 0 0]{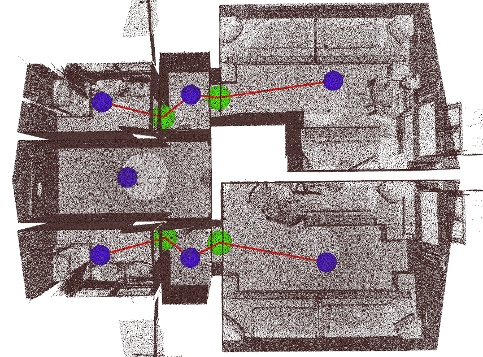}}
    		\end{minipage}  \\ \hline
    	\end{tabular}
    \end{table*}	
    
    \begin{table*}[h!]
    	\centering
    	\begin{tabular}{ | c | m{4.5cm} | m{4.4cm} | m{4.2cm} | }
    		\hline
    		& \makecell{Dataset 4} & \makecell{Dataset 5} & \makecell{Dataset 6}\\ \hline

    		\makecell{ \textbf{Input}  \\ points \\ boundary (x/y/z) \\ regions }
    		& \makecell{ 5,096,515 \\ 16.24m/13.90m/7.02m \\ 4}
    		& \makecell{ 291,701 \\ 26.46m/24.79m/8.39m \\ 3} 
    		& \makecell{ 1,199,532 \\ 24.15m/16.03m/3.37m \\ 11} \\ \hline
    		
    		\makecell{ \textbf{Info} \\ Voxel size \\ Time (seconds) \\ $a_{th}$ ($m^3$)} 
    		& \makecell{ 0.10 \\ 0.48 \\ 20 } 
    		& \makecell{ 0.15 \\ 0.27 \\ 20 } 
    		& \makecell{ 0.15 \\ 0.31 \\ 20 } \\ \hline
    		
    		\makecell[c]{  \\ \\ \\ \\ \textbf{Input} \\ \\ \\ \\ \\ \textbf{3D result}  \\ \\ \\ \\ \\ \textbf{1D result} }
    		& 
    		\begin{minipage}{.23\textwidth}
    			\centering
    			\subfloat{\includegraphics[width=0.6\linewidth]{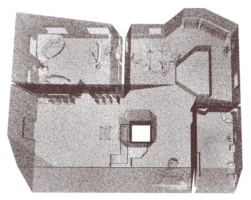}} \vfill
    			\subfloat{\includegraphics[width=0.6\linewidth]{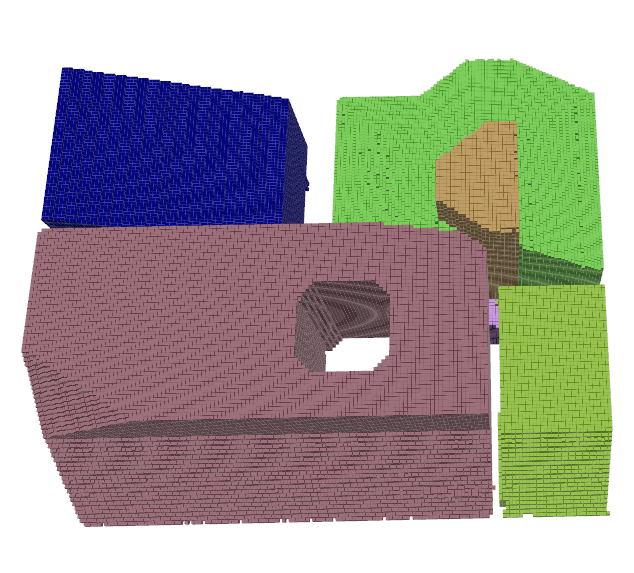}} \vfill
    			\subfloat{\includegraphics[width=0.6\linewidth]{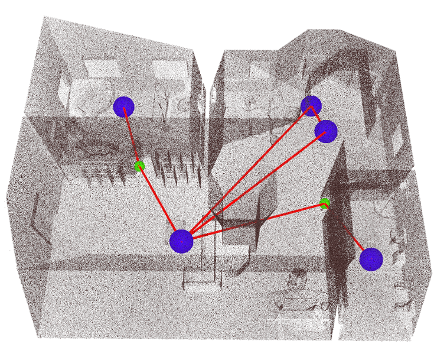}}	
    		\end{minipage}
    		& 
    		\begin{minipage}{.23\textwidth}
    			\centering
    			\subfloat{\includegraphics[width=0.6\linewidth]{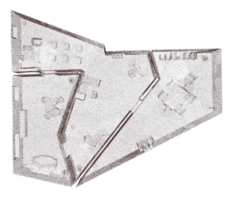}} \vfill	
    			\subfloat{\includegraphics[width=0.6\linewidth]{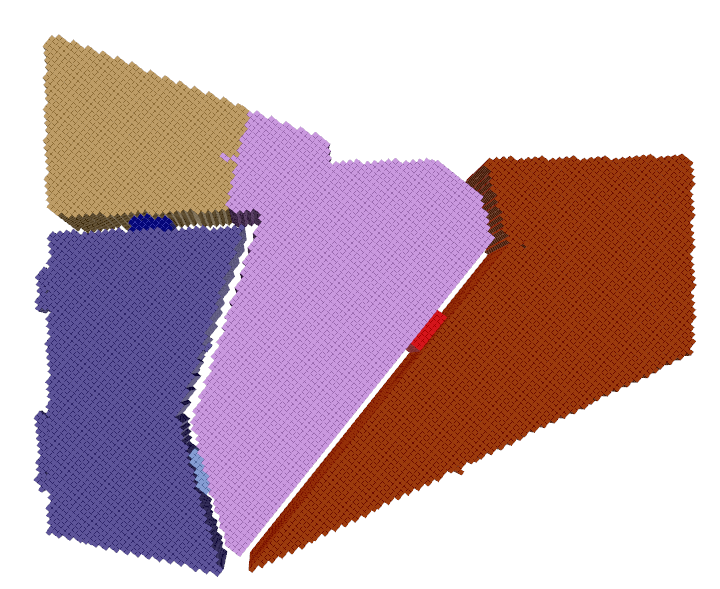}} \vfill
    			\subfloat{\includegraphics[width=0.6\linewidth]{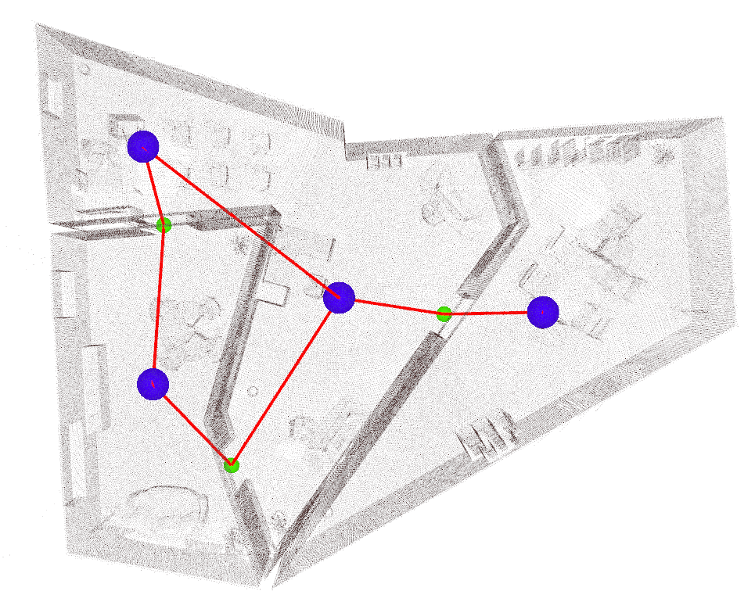}}
    		\end{minipage}
    		&
    		\begin{minipage}{.23\textwidth}
    			\centering
    			\subfloat{\includegraphics[width=0.7\linewidth]{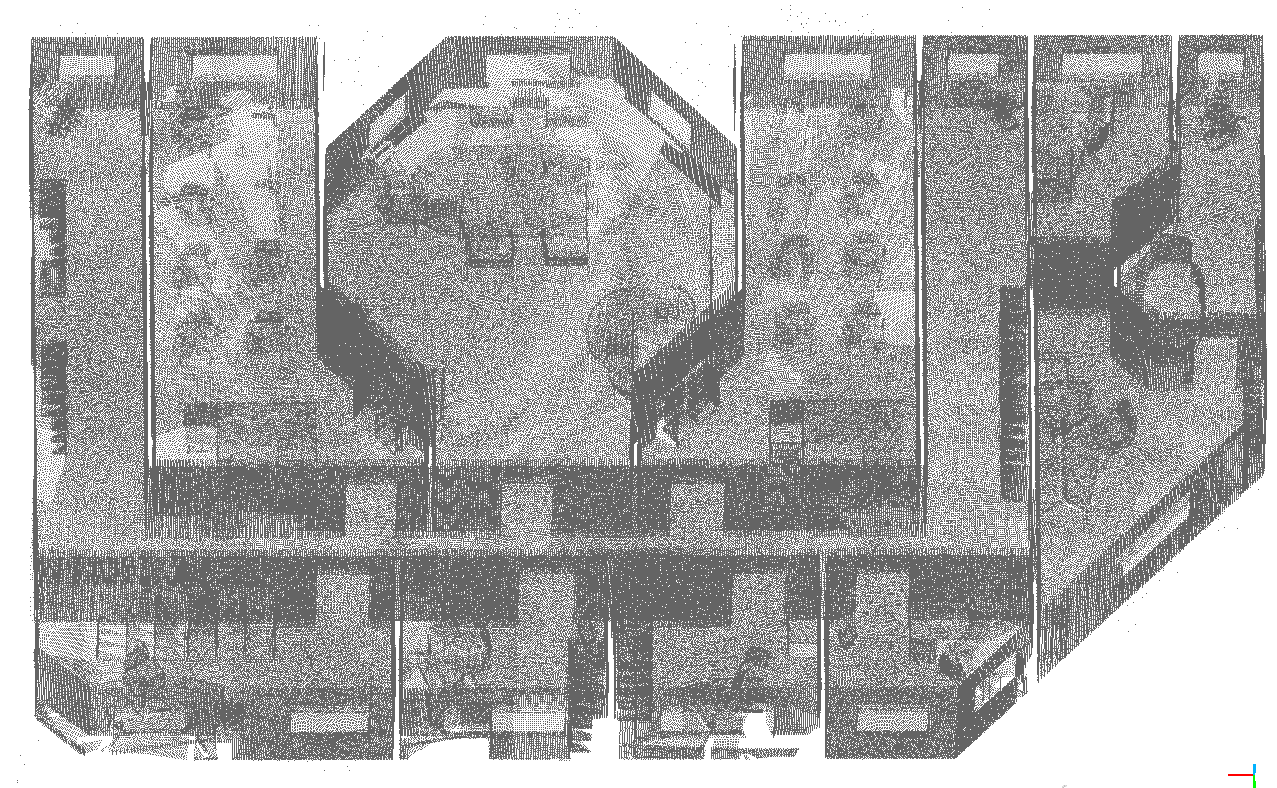}} \vfill	
    			\subfloat{\includegraphics[width=0.7\linewidth, trim=0 0 0 -100]{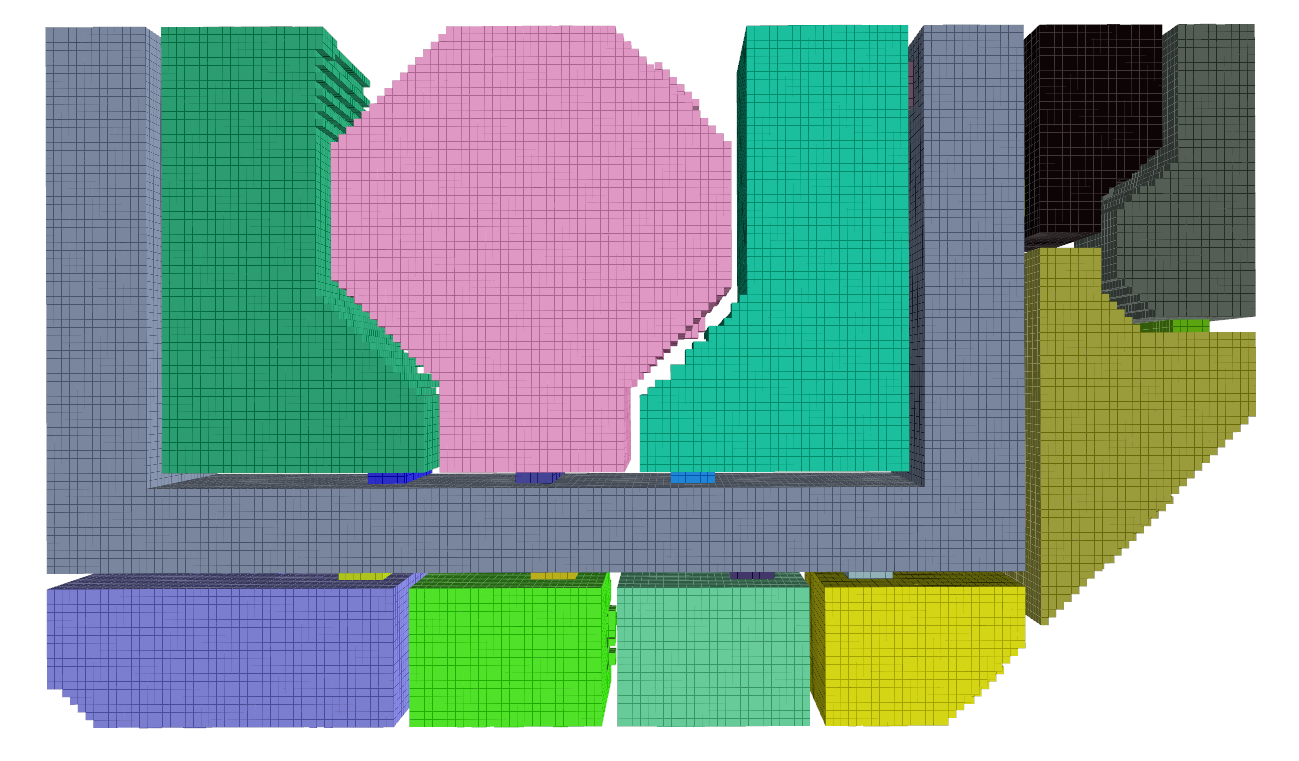}} \vfill
    			\subfloat{\includegraphics[width=0.7\linewidth, trim=0 0 0 -70]{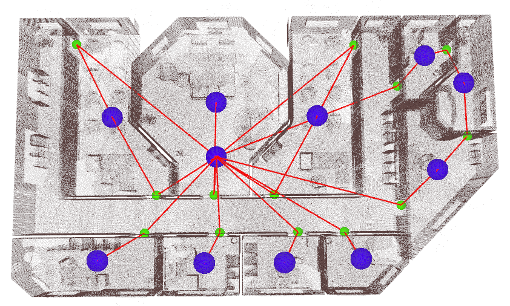}}
    		\end{minipage}  \\ \hline
    	\end{tabular}
    	\caption{Evaluation results on different datasets.}
    	\label{table:performance}
    \end{table*}		
    
    \begin{table*}[h!]
    	\centering
    	\begin{tabular}{ | m{3cm} | m{3cm} | m{3cm} | m{3cm} | m{3cm} | }
    		
    		\hline
    		\makecell{ \textbf{2D Input} \\ (Sliced from \\ the point cloud) } 
    		& \makecell{ \textbf{2D method} \\ (Area Graph) } 
    		& \makecell{ \textbf{2D method} \\ (MAORIS) } 
    		& \makecell{ \textbf{Ours (2D result)} \\ (Projection from \\ 3D clustering result \\ in vertical direction) } 
    		& \makecell{ \textbf{2D Ground Truth} } \\ \hline
    		
    		\centering
    		\includegraphics[width=0.85\linewidth]{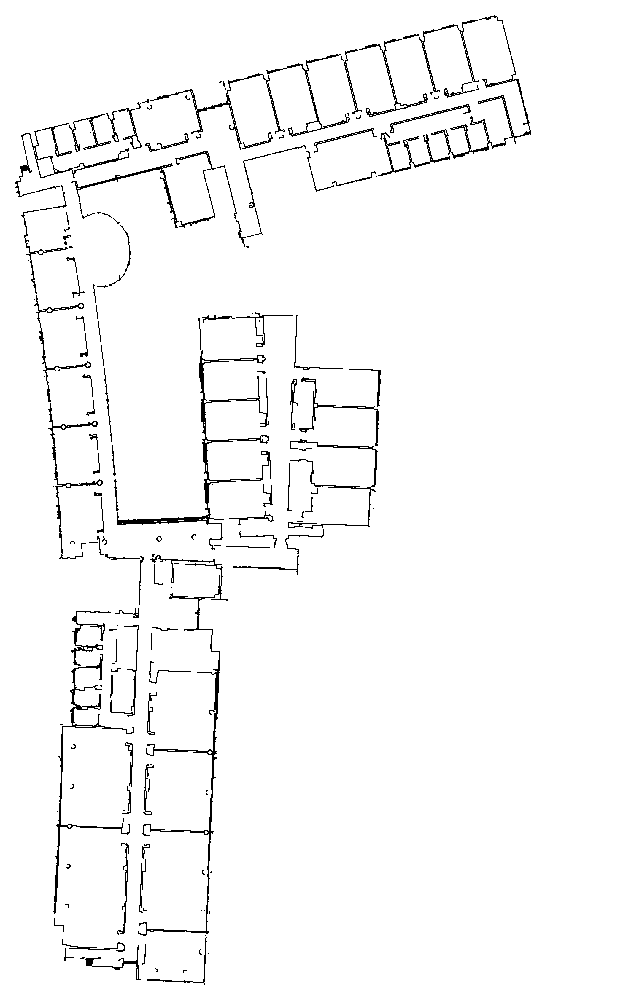}
    		& 
    		\centering
    		\includegraphics[width=0.85\linewidth]{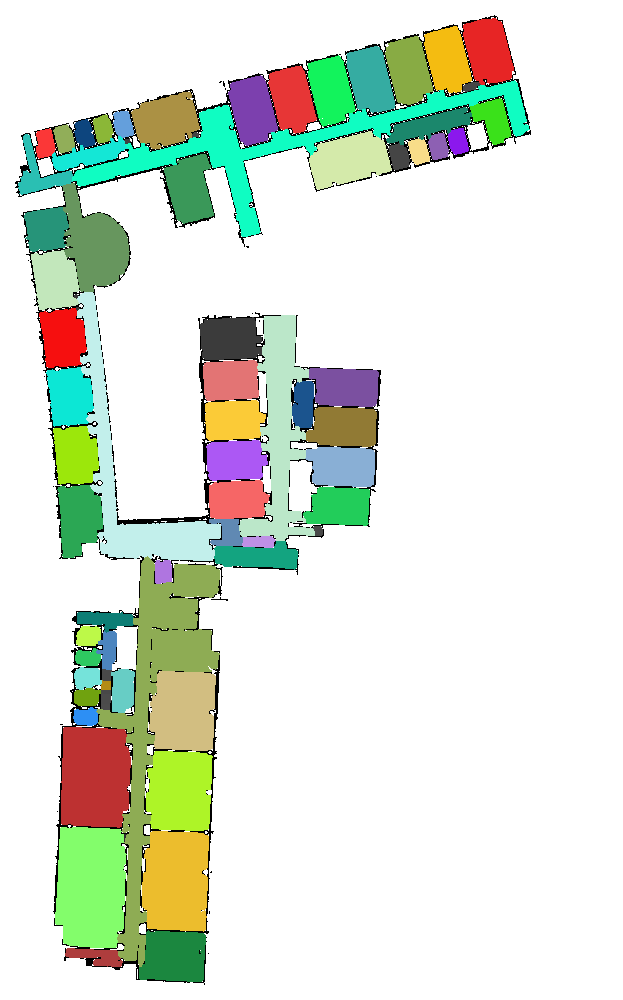}
    		& 
    		\centering
    		\includegraphics[width=0.85\linewidth]{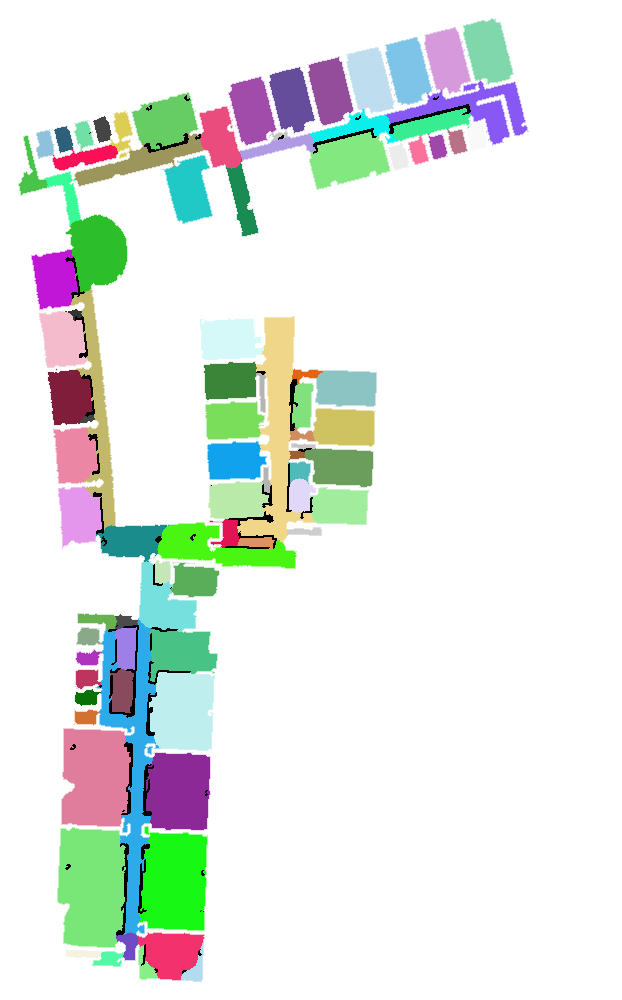}
    		& 
    		\centering
    		\includegraphics[width=0.85\linewidth]{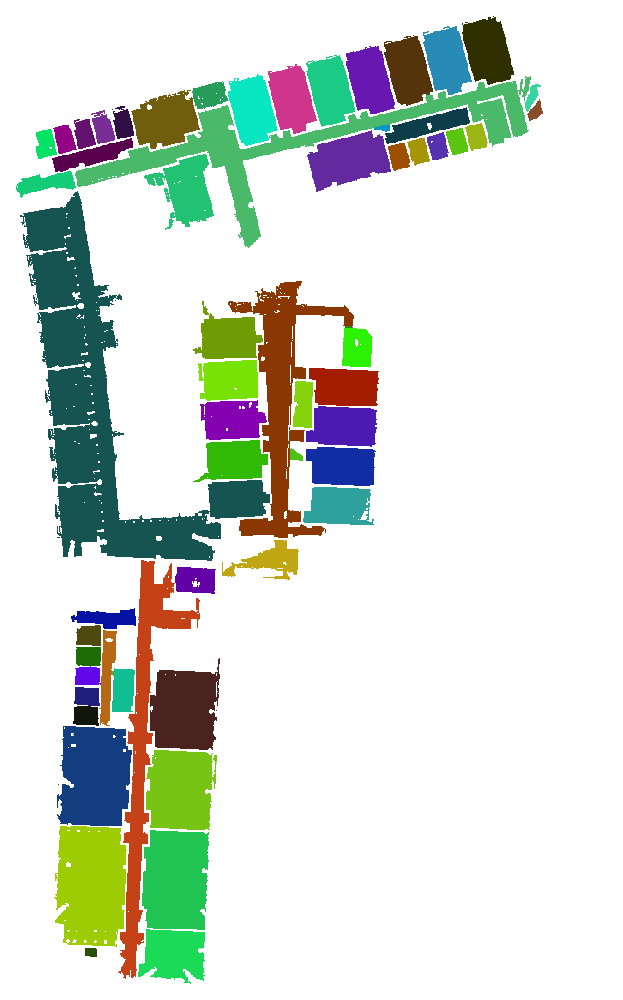}
    		&
    		\begin{center}
    			\includegraphics[width=0.85\linewidth]{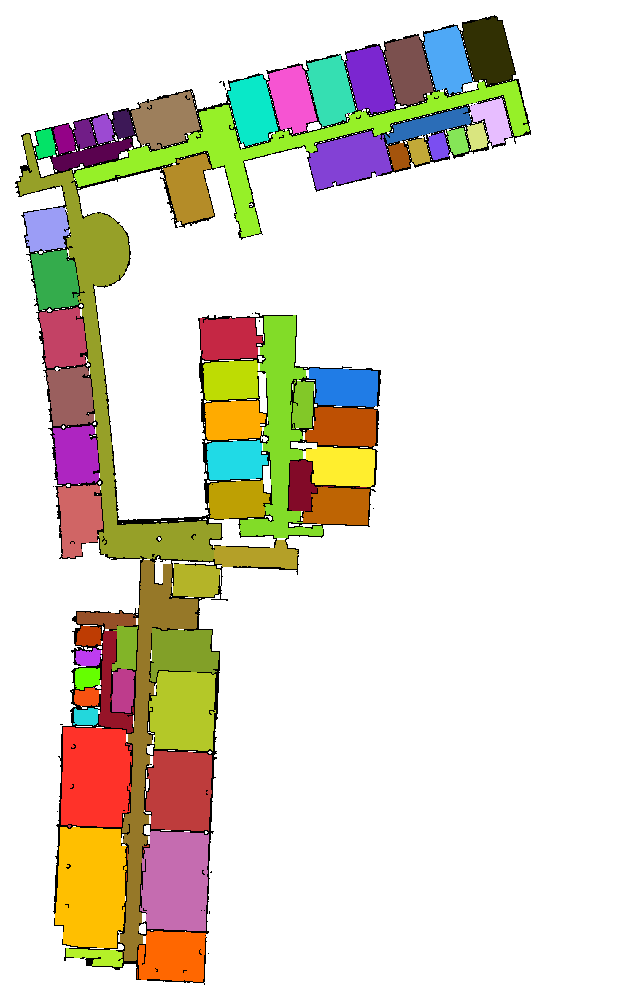}
    		\end{center}
    		\\ \hline
    		
    		\centering
    		\includegraphics[width=0.55\linewidth]{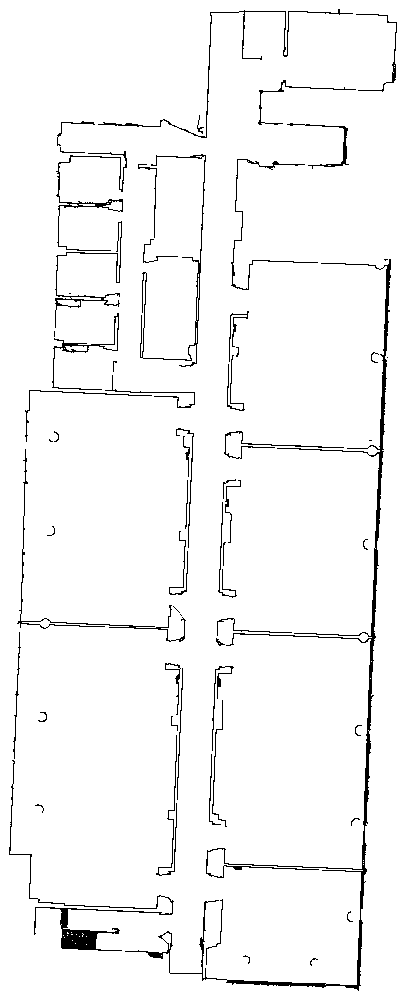}
    		& 
    		\centering
    		\includegraphics[width=0.55\linewidth]{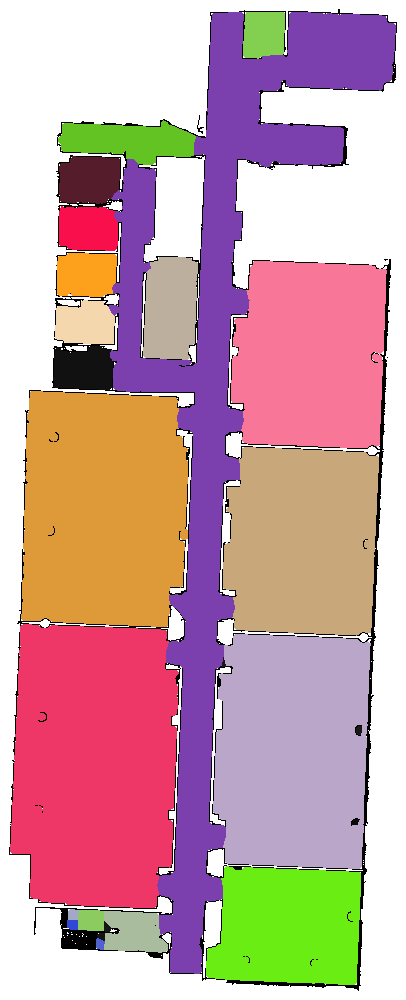}
    		& 
    		\centering
    		\includegraphics[width=0.55\linewidth]{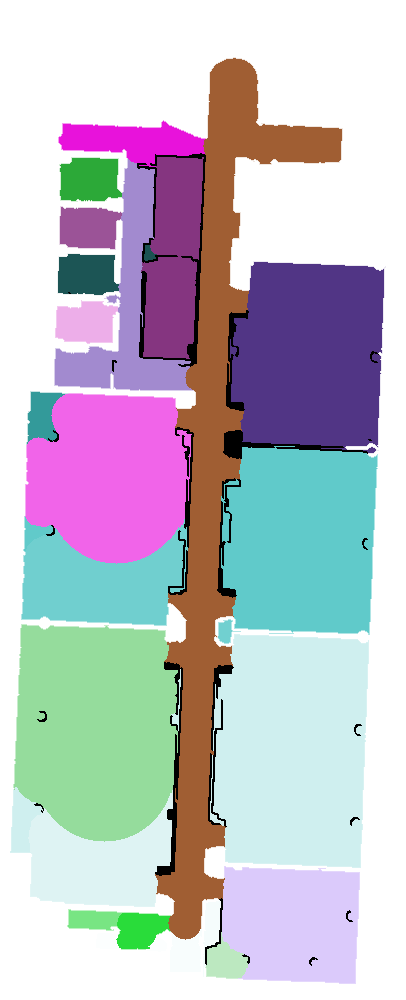}
    		& 
    		\centering
    		\includegraphics[width=0.55\linewidth]{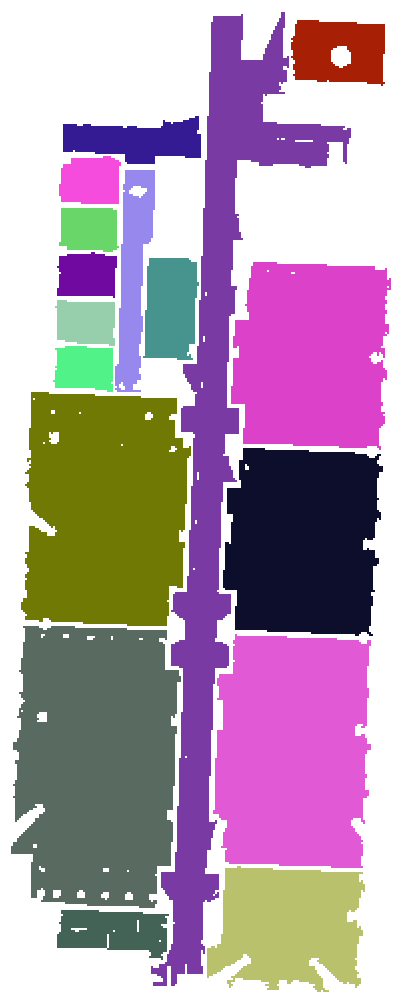}
    		&
    		\begin{center}
    			\includegraphics[width=0.55\linewidth]{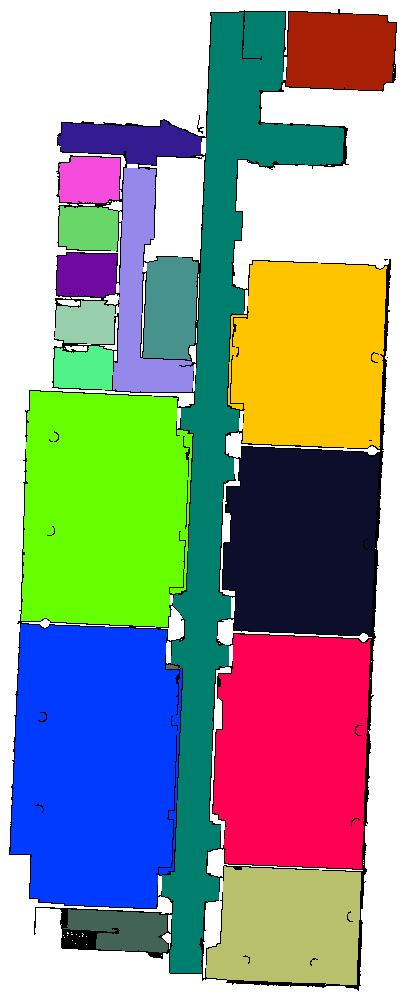}
    		\end{center}
    		\\ \hline		
    		
    		\centering
    		\includegraphics[width=0.55\linewidth]{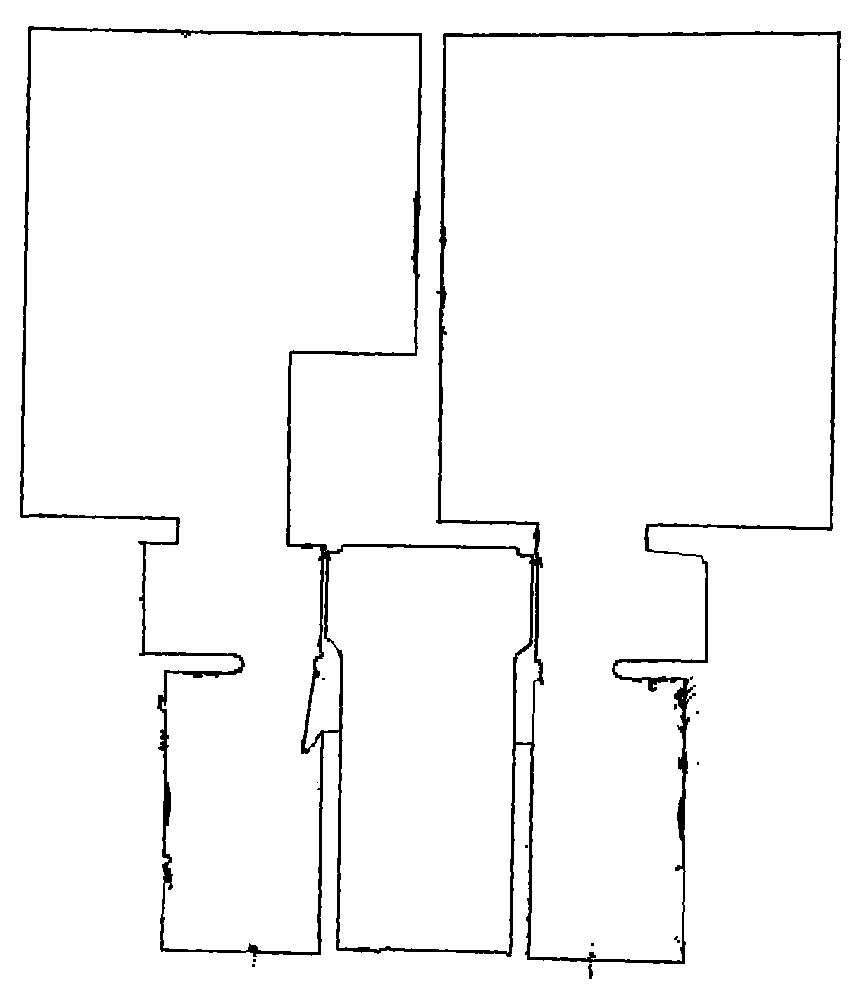}
    		& 
    		\centering
    		\includegraphics[width=0.55\linewidth]{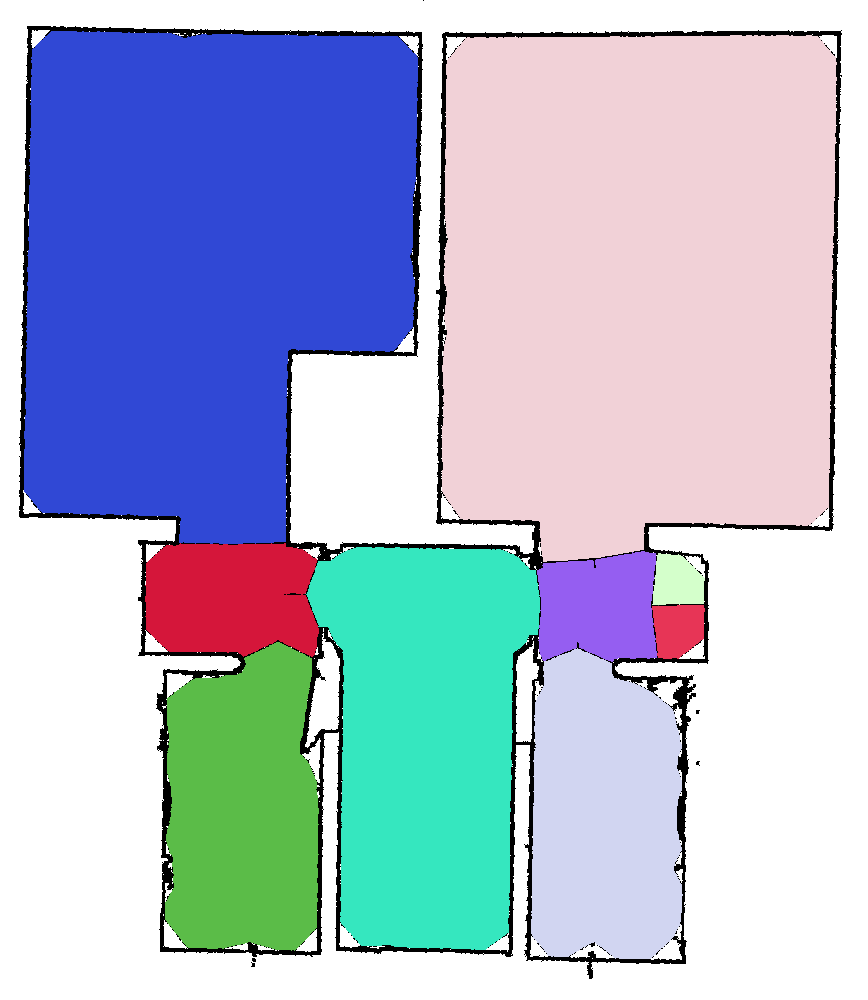}
    		& 
    		\centering
    		\includegraphics[width=0.55\linewidth]{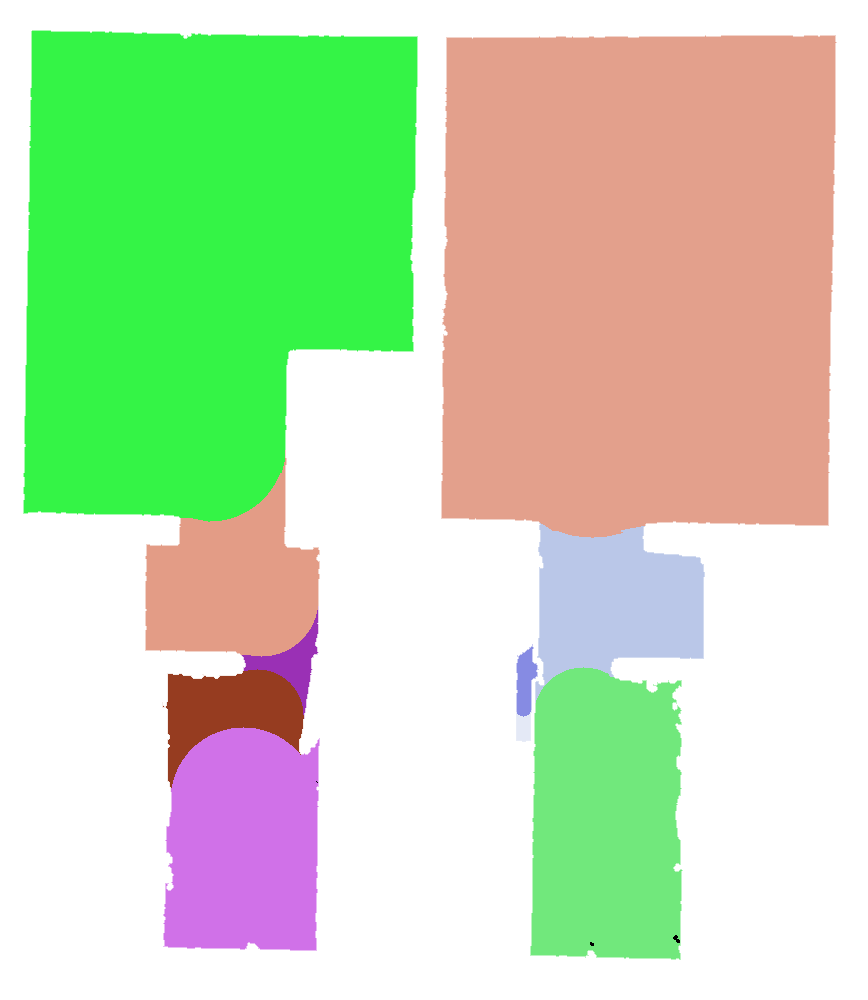}
    		& 
    		\centering
    		\includegraphics[width=0.55\linewidth]{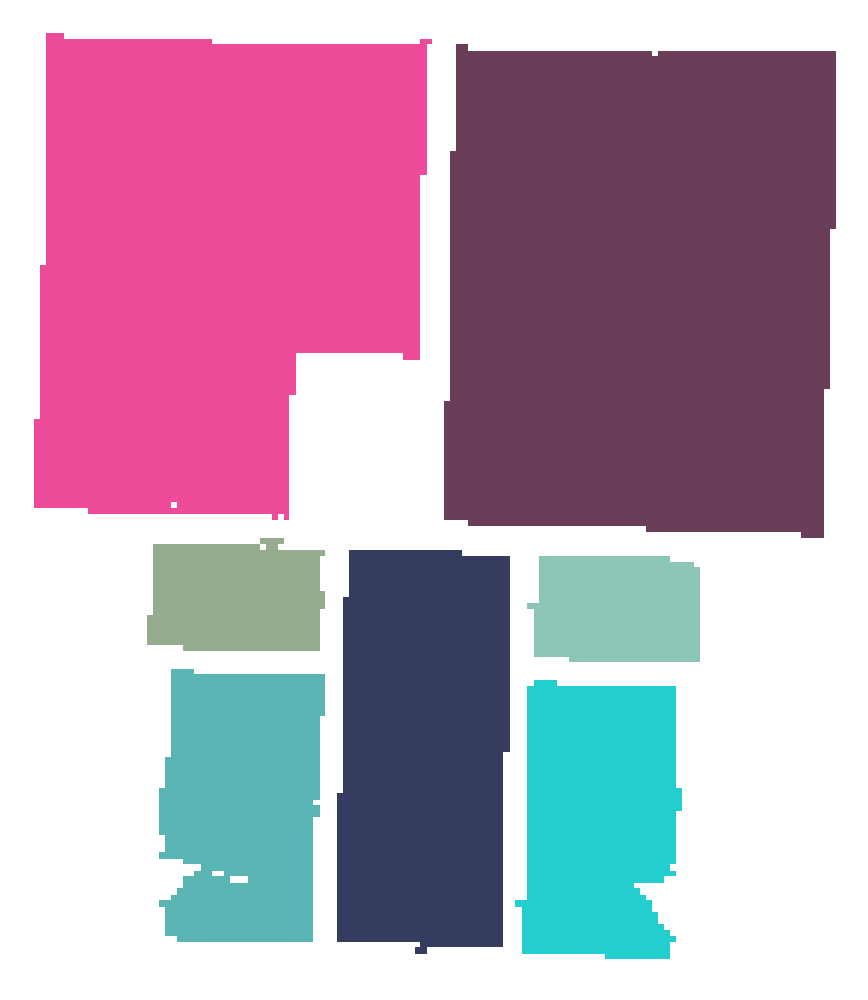}
    		&
    		\begin{center}
    			\includegraphics[width=0.55\linewidth]{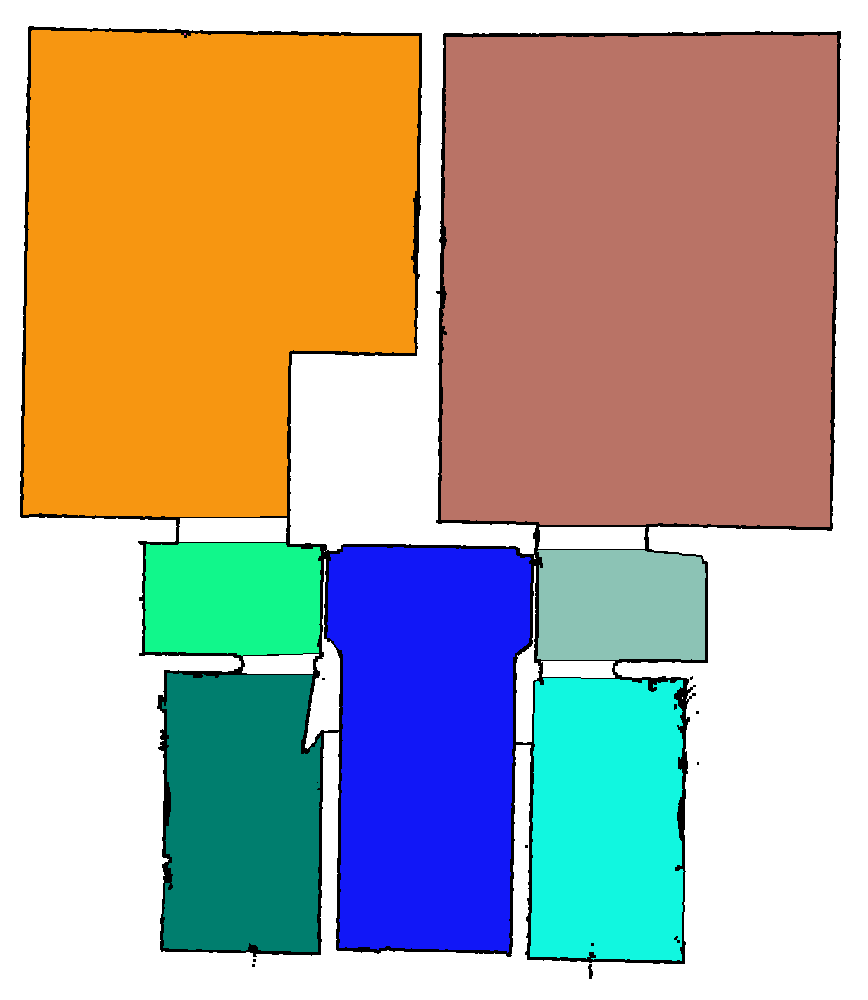}
    		\end{center}
    		\\ \hline	
    		
    		\centering
    		\includegraphics[width=0.55\linewidth]{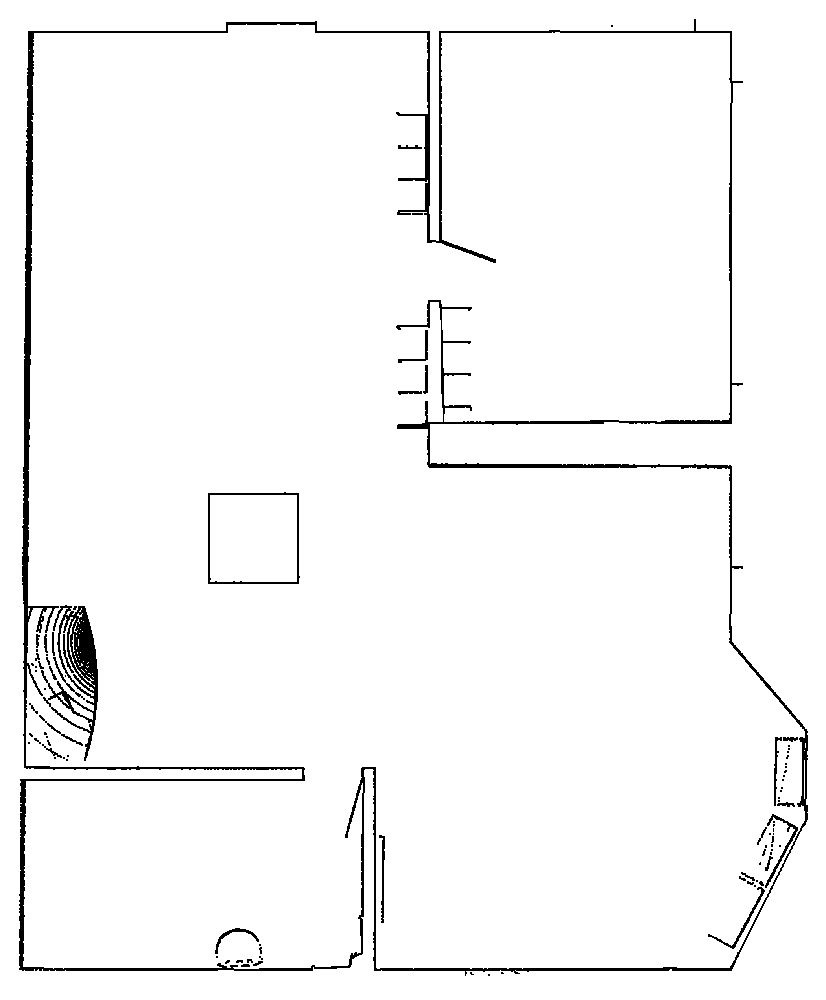}
    		& 
    		\centering
    		\includegraphics[width=0.55\linewidth]{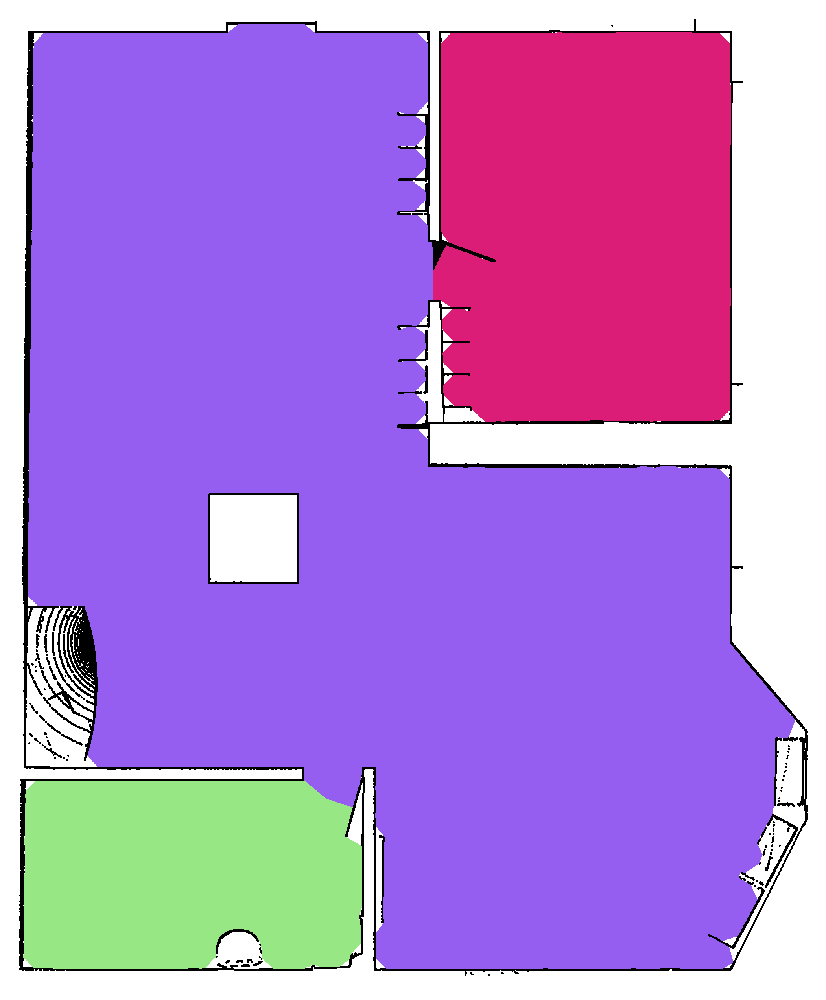}
    		& 
    		\centering
    		\includegraphics[width=0.55\linewidth]{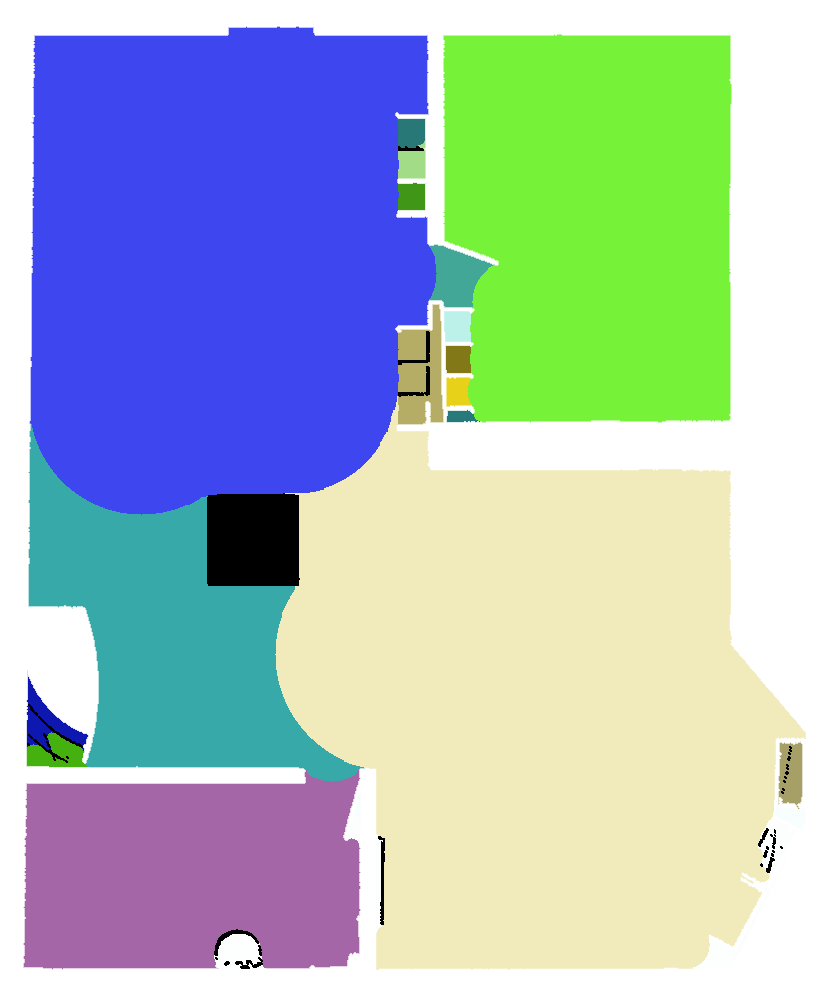}
    		& 
    		\centering
    		\includegraphics[width=0.55\linewidth]{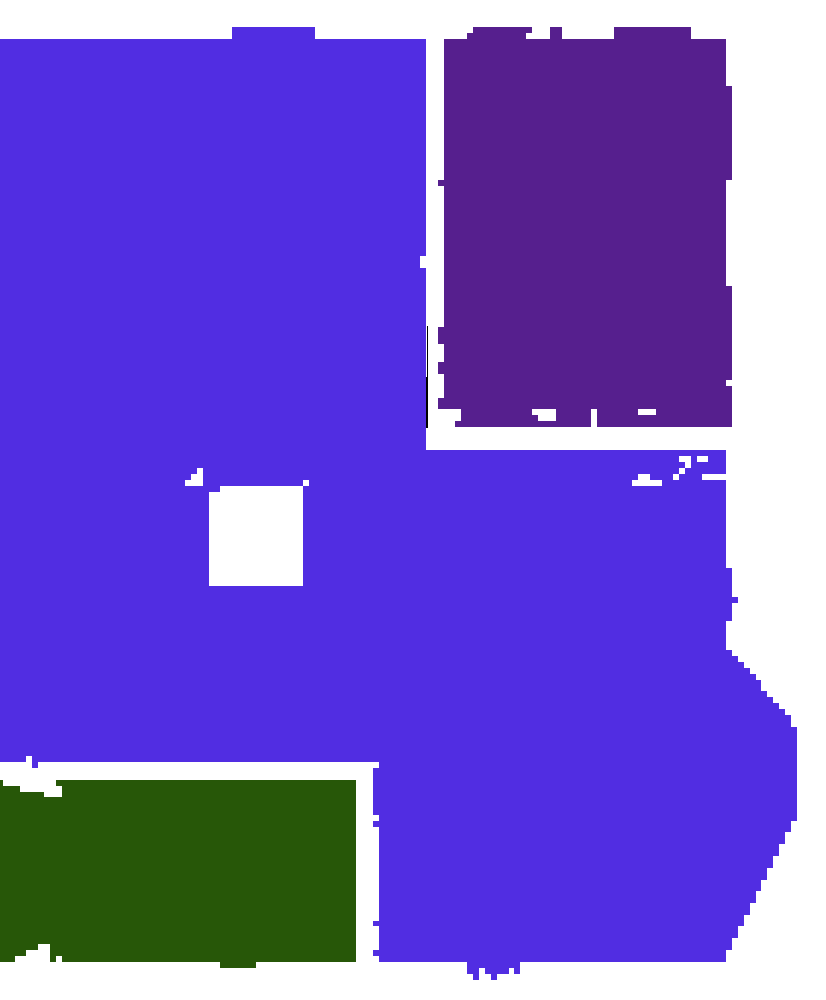}
    		&
    		\begin{center}
    			\includegraphics[width=0.55\linewidth]{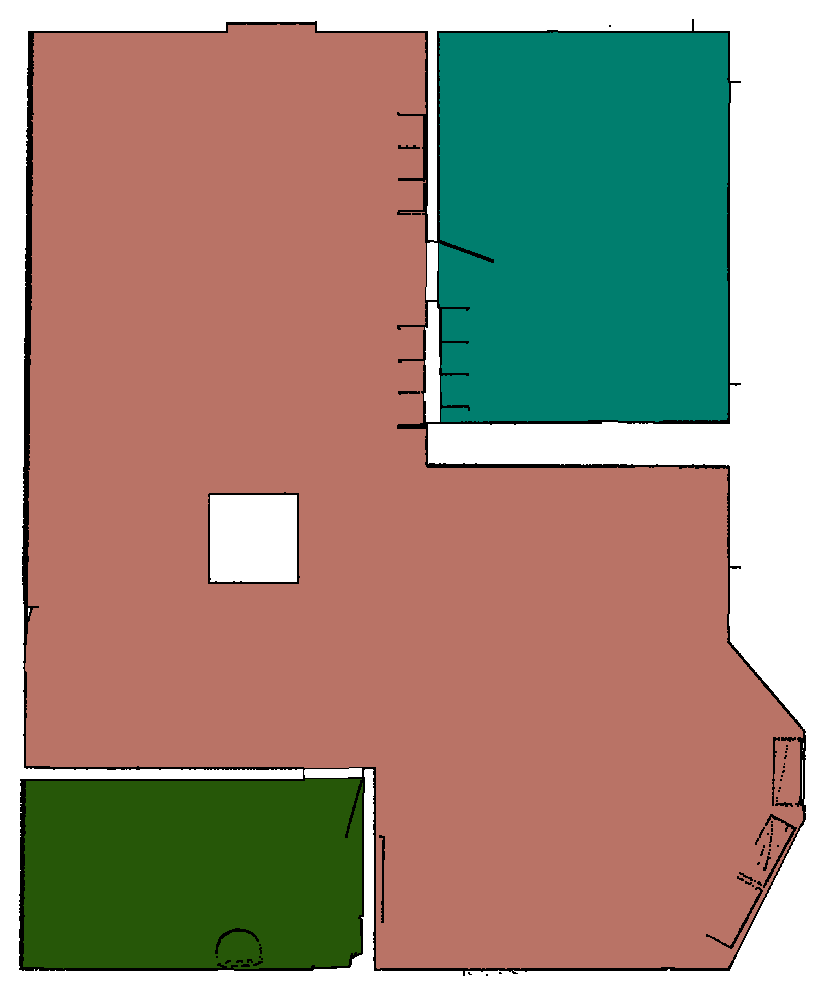}
    		\end{center}
    		\\ \hline	
    		
    		\centering
    		\includegraphics[width=0.55\linewidth]{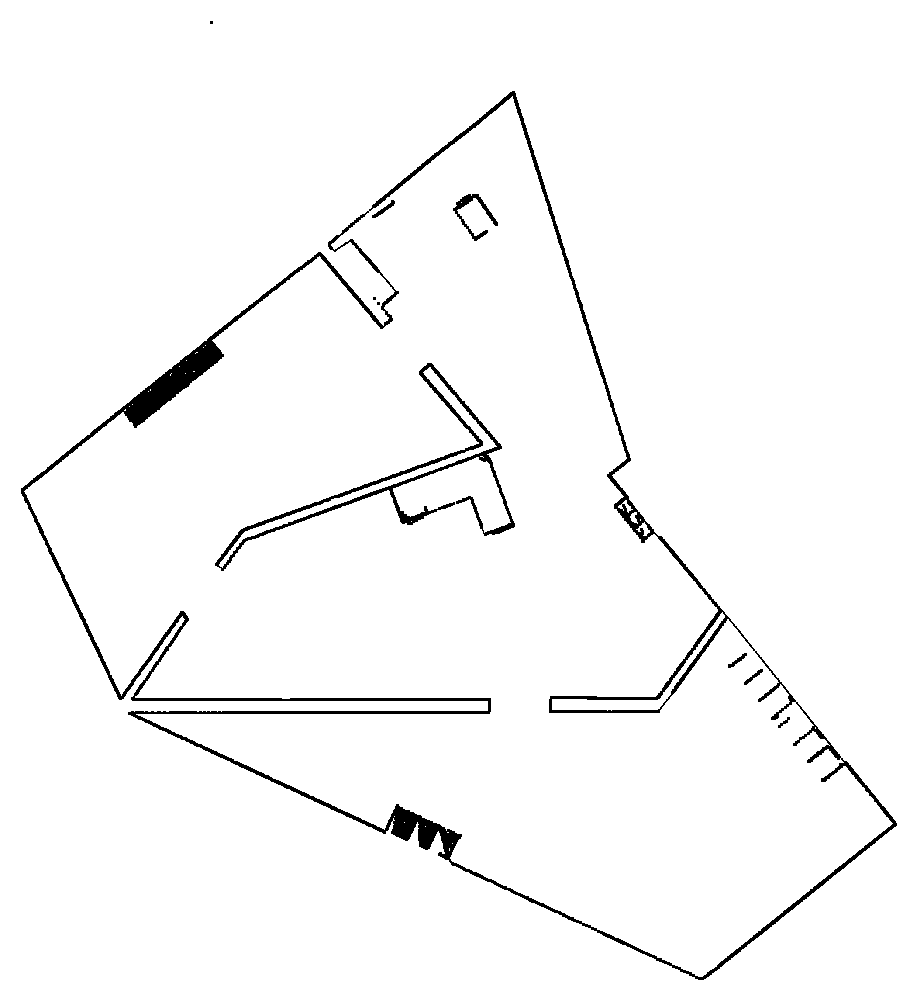}
    		& 
    		\centering
    		\includegraphics[width=0.55\linewidth]{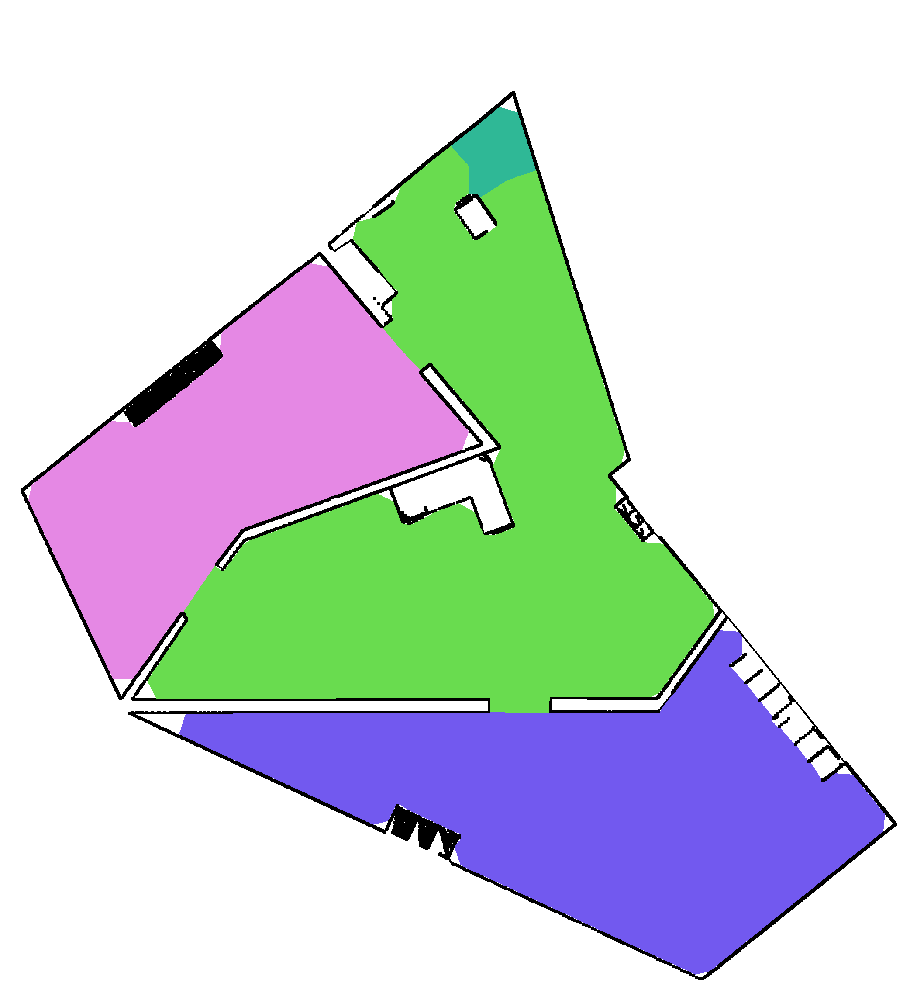}
    		& 
    		\centering
    		\includegraphics[width=0.55\linewidth]{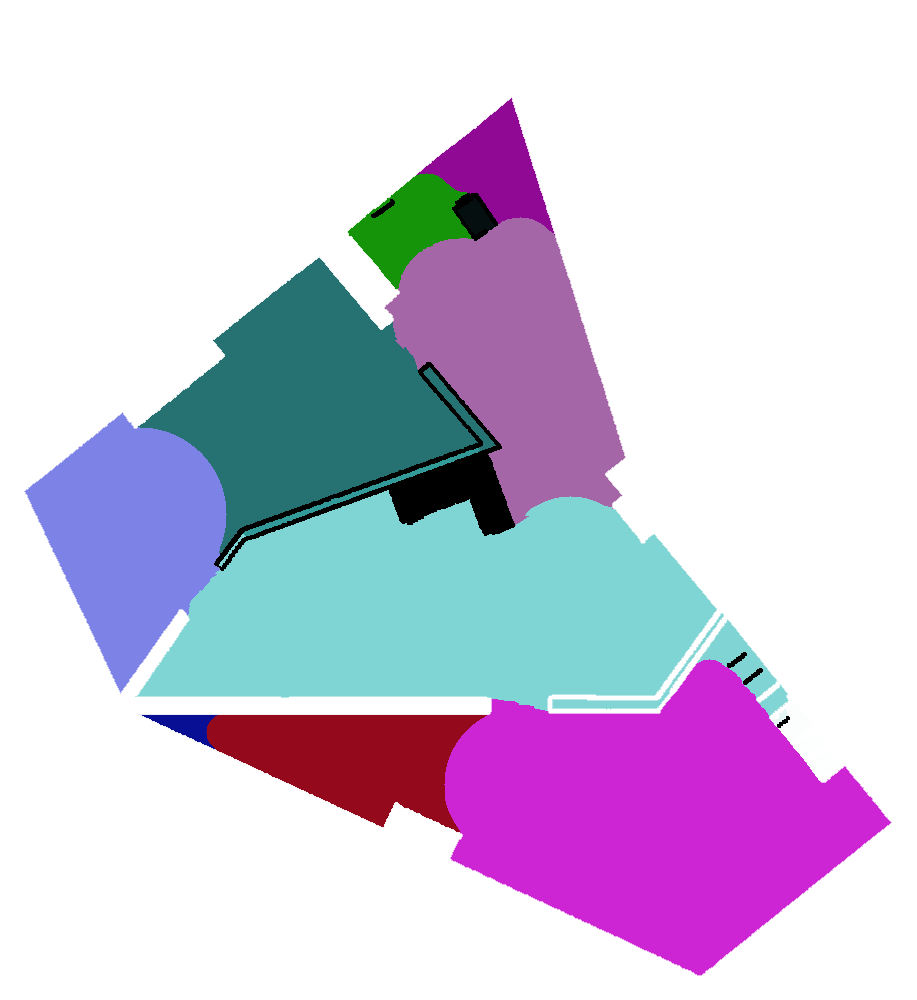}
    		& 
    		\centering
    		\includegraphics[width=0.55\linewidth]{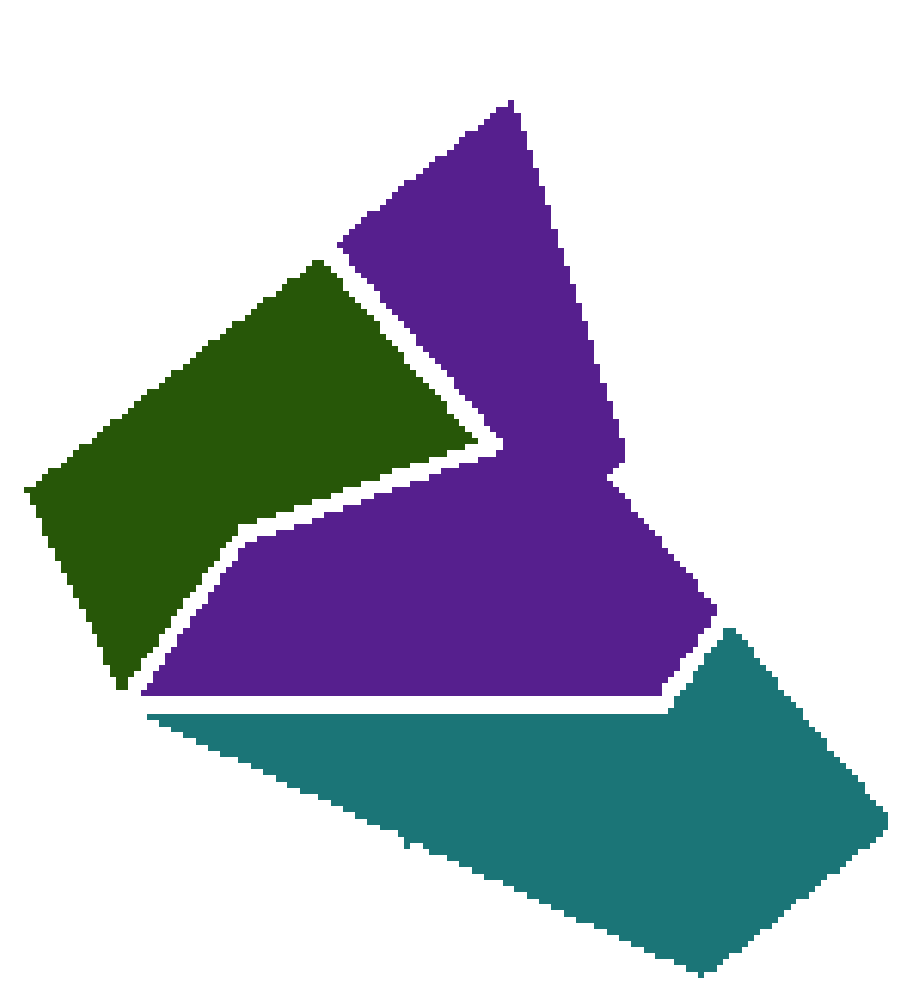}
    		&
    		\begin{center}
    			\includegraphics[width=0.55\linewidth]{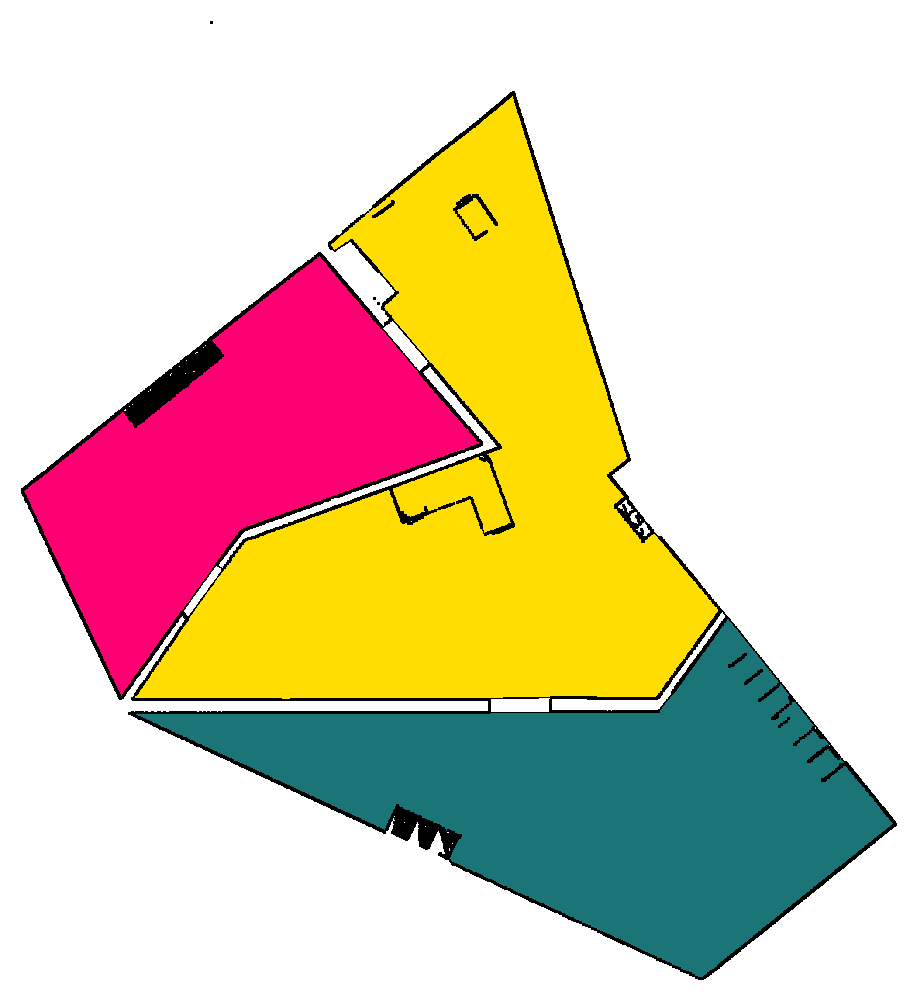}
    		\end{center}
    		\\ \hline		
    		
    		\centering
    		\includegraphics[width=0.55\linewidth]{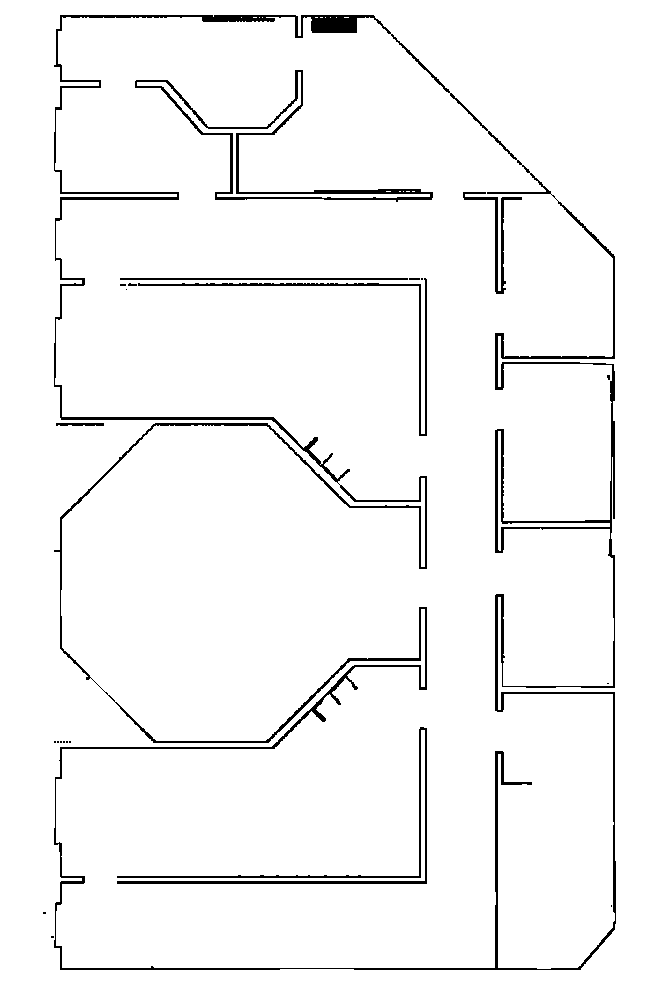}
    		& 
    		\centering
    		\includegraphics[width=0.55\linewidth]{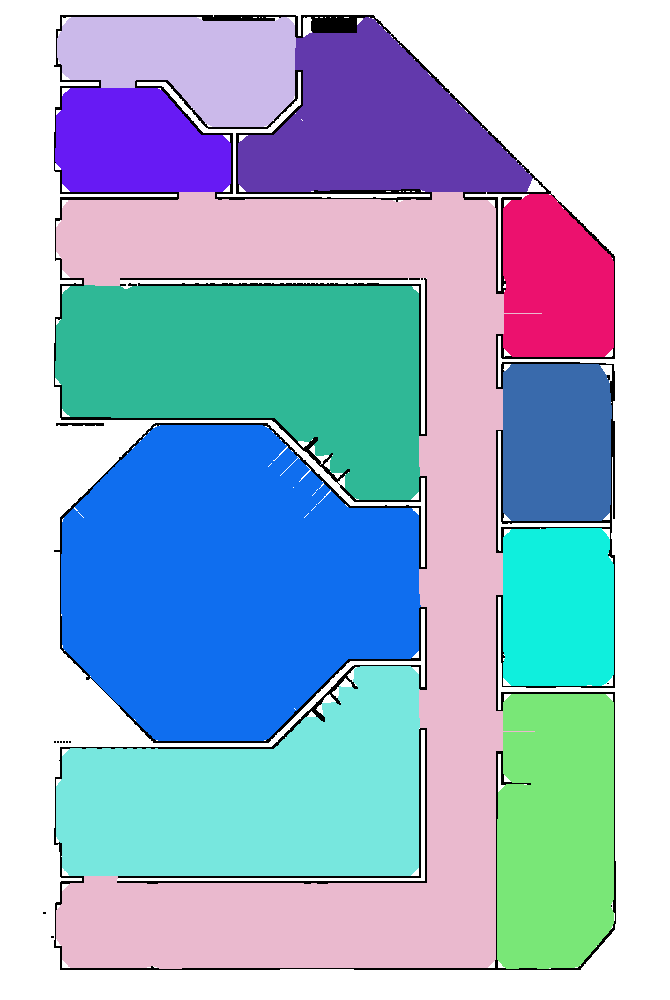}
    		& 
    		\centering
    		\includegraphics[width=0.55\linewidth]{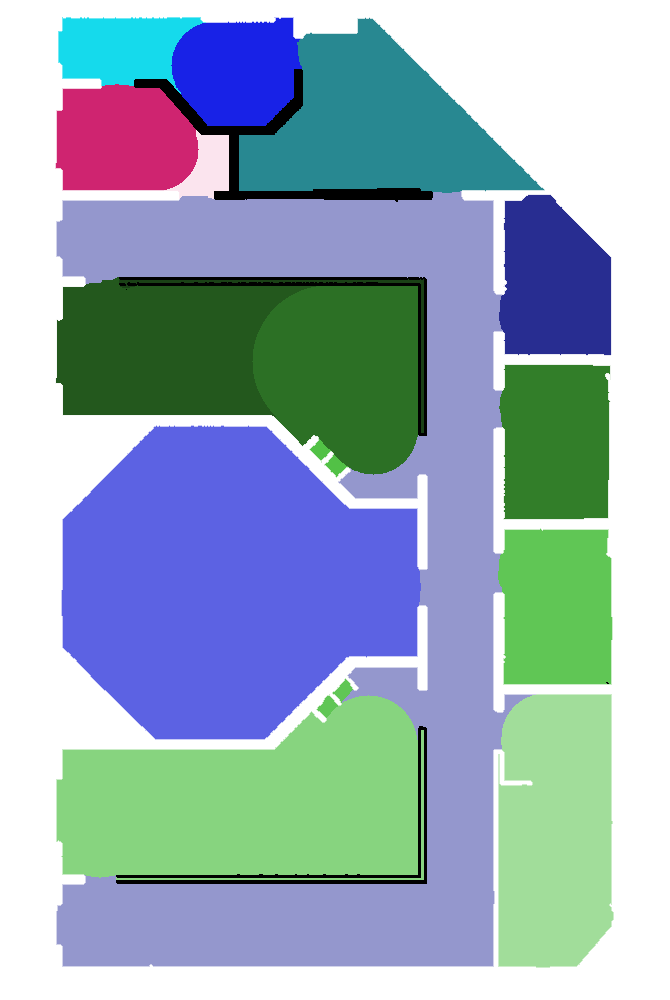}
    		& 
    		\centering
    		\includegraphics[width=0.55\linewidth]{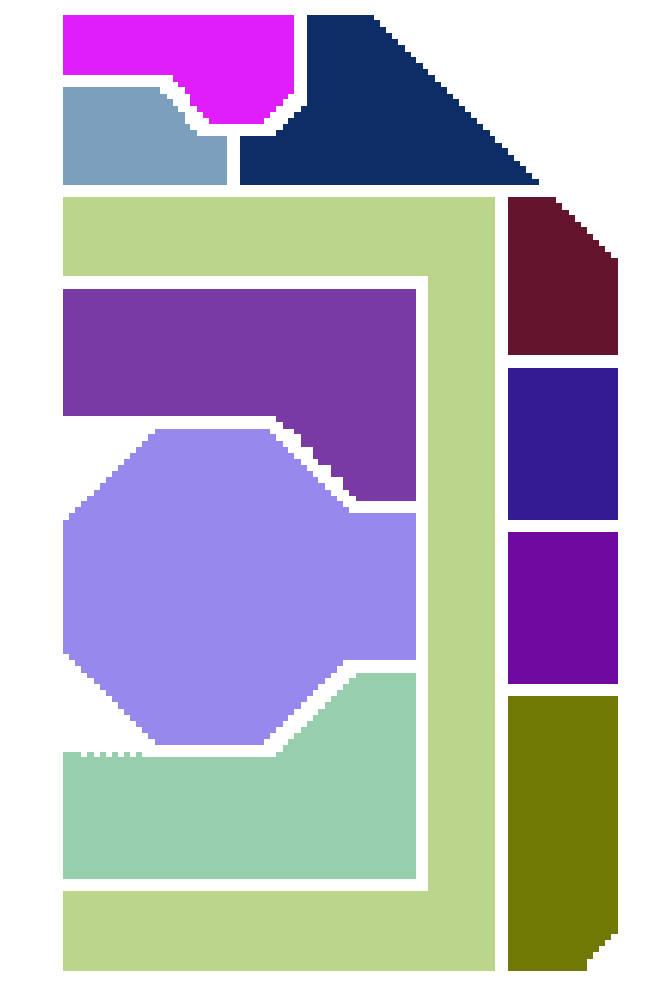}
    		&
    		\begin{center}
    			\includegraphics[width=0.55\linewidth]{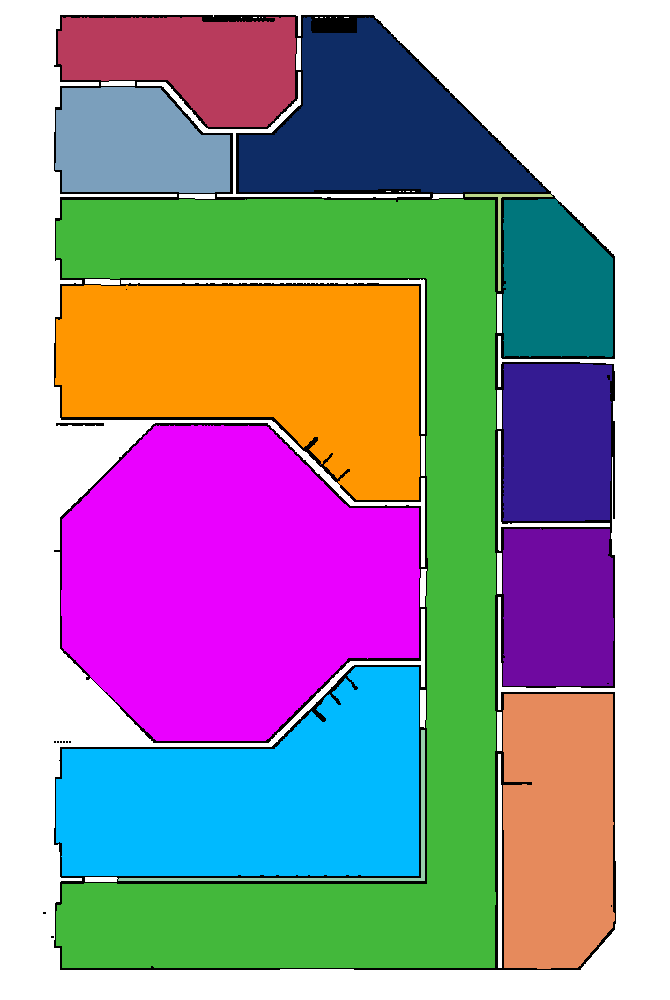}
    		\end{center}
    		\\ \hline							
    		
    	\end{tabular}
    	\caption{Comparisons with state-of-the-art 2D segmentation results.}
    	\label{table:area}
    \end{table*}		
	
\section{Evaluation}
	\label{sec:ex}
    Section \ref{subsec:performance} evaluates the performance and quality of our multi-dimensional topological maps on a variety of datasets. Section \ref{subsec:comparisons} compares our two-dimensional projection $region$ segmentation result with existing methods and ground truth. Section \ref{subsec:limitation} discusses the limitations of our algorithm.
    
    {\it Datasets}: We use a variety of real-world point clouds for our evaluation. Table \ref{table:performance} shows six single storey point clouds generated by terrestrial laser scanners.\\ Dataset 1 and Dataset 2 were provided by our Mobile Autonomous Robotic Systems Lab (MARS Lab) at Shanghai Tech University\footnote{\url{https://robotics.shanghaitech.edu.cn/datasets/3D_topo}}. It should be noted that those two datasets were collected before furniture was moved into the building. Nevertheless, it is also realistic to generate similar maps in buildings with furniture, as we showed in \citet{he2019furniture}, which uses the robot from \citet{chen2020advanced} to create furniture-free 3D point cloud maps. The other datasets are the open-source datasets published by the Visualization and MultiMedia Lab at the University of Zurich. They are collected from a real environment and feature objects such as furniture.
    
    {\it Platform:} For the experiments a single-threaded C++ implementation on an Intel i7-8750H (2.2G) CPU was used. To maintain the stability of the results, we set the CPU Turbo frequency to only 3.8Ghz.
	
\subsection{Performance in Different Datasets}
	\label{subsec:performance}
	Table \ref{table:performance} shows the input point cloud and the output topometric maps of different dimensions. Our algorithm works well on the test datasets. In particular, there is much noise in Dataset 1 due to reflections from glass, but our method is robust enough to to filter-out any outliers and obtains a meaningful 3D result. The Datasets 3 and 4 are buildings with slanted ceilings. In both datasets, rooms with slanted ceilings will still be identified as a single $volume$. In most cases, our topological graph shows functional connectivity and accessibility of $regions$ (e.g., Dataset 3, 4, 5, 6). A few cases of $region$ over-seg\-mentation in Dataset 1 can be observed. It was caused by 2D area graph segmentation and would not influence the accessibility of $regions$. 
	
	A few vertices have no edge in Dataset 1, 2 because the point cloud is incomplete. There are hollows above some doors, and then our algorithm was unable to generate the $column$ in this position and identify those doors. Also note that Dataset 2 is not part of Dataset 1. 
	
	The input of our method is a point cloud map of the entire building, so real-time performance is not required. But our algorithm runs within 5 seconds on the test datasets. There are two factors that affect the run time of the algorithm, the building size and the selected voxel size. It can be seen from the Dataset 1 - 6 of Table \ref{table:performance} that the run time of the algorithm is proportional to the size of the building. At the same time, the building size of Dataset 4 is smaller, but the run time is longer than that of Dataset 5. This is because Dataset 4 selects a smaller voxel size.

\subsection{Evaluation of Clustering Result in the Vertical Direction}
	\label{subsec:comparisons}
	Our algorithm combines a clustering method in the vertical direction (Section \ref{method:region}) and a space partition method in the horizontal direction (Section \ref{method:area}). It is very difficult to generate 3D ground truth segmentation data, so we evaluate our 3D segmentation by comparing its 2D projection to 2D segmentation algorithms. Table \ref{table:area} shows that our clustering method in the vertical direction with the comparison methods in an indoor environment. We compare with two recent 2D space partitioning methods. The work of \citet{mielle2018method} divides navigation maps into a semantic representation. Their method compares with many traditional works and achieves a state-of-the-art (SotA) result. The work of \citet{hou2019area} is a recent approach and is suitable for a large scale environment. It is superior to SotA approaches in complex indoor environments with well-defined parameters.
	
	\markCM{ The first column in Table \ref{table:area} is the input of Area Graph segmentation algorithm \citet{hou2019area} and MAORIS algorithm, \citet{mielle2018method}. The 2D grid map is sliced from the 3D point cloud. Since the latter method is only applicable to environments without furniture, we manually filter the noise in the 2D grid map as input to the relevant 2D segmentation method. This only benefits the performance of the two 2D methods, while our approach uses the noisy 3D point cloud as input. The second column shows the segmentation results  of the Area Graph algorithm. The third column shows the results generated by the MAORIS algorithm. In the fourth column, we project our 3D $region$ result to 2D without 2D horizontal direction space partitioning method. The ground truth is made by hand according to the input image, the painting process is as follows: Refer to the 3D point cloud and divide the map where the door exists (stained in different colors). All the point clouds we use can be found online. }
	

	\markCM{We use Matthew’s correlation coefficient (MCC) introduced by \citet{mielle2018method} for segmentation measurement against the human-made ground truth. One advantage of the MCC for evaluating map segmentation is that it stays balanced even when the classes are of different sizes. In MCC, each segmented color block (region) is associated with the ground truth color block. The true positive (tp), false positive (fp), false negative (fn) and true negative (tn) are defined as follows:
	\begin{itemize}
		\item tp: number of pixels in both the segmented and the ground truth regions
		\item fp: number of pixels in the segmented but not in the ground truth regions
		\item fn: number of pixels not in the segmented but in the ground truth regions
		\item tn: number of pixels in neither the segmented nor the ground truth regions
	\end{itemize}
	$$	MCC = \frac{tp * tn - fp * fn}{\sqrt{(tp +fp)(tp +fn)(tn + fp)(tn + fn)}}	$$
	The MCC ranges between -1 and 1. The best result is 1, 0 is no better than guessing, and -1 is indicating total disagreement.} 

	\begin{table}[]
		\centering
		\begin{tabular}{@{}cccc@{}}
			\toprule
			MCC       & Area Graph  & MAORIS & Ours (Projection)  \\ \midrule
			Dataset 1 & \textbf{0.606}  & 0.498    & {0.532} \\
			Dataset 2 & 0.589  & 0.615    & \textbf{0.976} \\
			Dataset 3 & 0.783  & 0.840    & \textbf{0.993} \\
			Dataset 4 & 0.747  & 0.279    & \textbf{0.975} \\
			Dataset 5 & 0.781  & 0.377    & \textbf{0.997} \\
			Dataset 6 & 0.930  & 0.646    & \textbf{0.993} \\ \bottomrule
		\end{tabular}
		\caption{The MCC score of different method.}
		\label{table:mcc}
	\end{table}

	\markCM{It can be seen from Table \ref{table:mcc} that our algorithm can achieve excellent results on most test data, better than Area Graph segmentation algorithm and MAORIS algorithm. However, our algorithm scores relatively low on Dataset 1. This is because Dataset 1 has many glass walls (e.g. Part 1 in Fig. \ref{fig:area}) and LiDAR cannot obtain the corresponding point cloud. Since Area Graph segmentation algorithm and MAORIS algorithm work in two dimensions, they are not affected by these glass walls.}
	
	By comparing the current SotA segmentation algorithms based on the horizontal direction, our clustering method based on the vertical direction shows superiority. As shown in the result, doors, pillars or irregular room shapes can lead to an over-segmentation in the horizontal segmentation. At the same time, our method shows a good result similar to the ground truth. 
	
\subsection{Limitations}
	\label{subsec:limitation}
	There are mainly three limitations in this work. 
	One technical limitation of our current implementation is the continuity of ceiling height. We merge $columns$. In \ref{method:volume} we generate $volumes$ by merging adjacent $columns$. The maximum allowed height difference between different $columns$ is $10\%$ of the length of this $column$. According to this height constraint, we can deal with the uneven ceilings in some cases, as shown in Dataset 4. Significant changes in local ceiling height will lead to segmentation. This height constraint applies to most situations, such as the partition from door to room. But in other cases, such as chandeliers and pipes on the ceiling, it may cause incorrect segmentation.
	
	Another limitation is the quality of the dataset. Incomplete scans can cause hollows in the point cloud, such as holes in the ceiling or ground. Holes that appear on the ground can be filled according to ground height calculated by \ref{method:storey}, but we are unable to deal with a hole that appears in the ceiling. It will result in missing $columns$. One example is shown in Dataset 2. There is a hole on the door of the rightmost (top) room. So there are no edges out of the rightmost (top) room, because of the missing $columns$. The glass in the environment can also affect the quality of a dataset. The problem caused by glass has two aspects, one is the transmission, and the other is reflection. The reflection of the laser will generate a mirror-symmetric entity about the glass in the point cloud. The transmission will cause the laser to penetrate the glass, which means that there is no point entity in the corresponding position of the point cloud, as shown in Fig. \ref{fig:area} (1). Area Graph segmentation mitigates the effects of this problem. 
	
    The last limitation is the voxel size. A large voxel size would fill the openings between walls. It also reduces the resolution of the results. However, a small voxel size will significantly increase the memory consumption, and make our approach sensitive to the incompleteness of the point cloud. Our algorithm does not save the voxel in the 3D voxel occupancy map; in the future, an incremental generation strategy can improve the efficiency of the algorithm. We could also employ an octree representation instead of a 3D grid to alleviate the size constraint to a certain extend.
    

\section{Conclusions}
    \label{sec:conclusion}
    This paper presents a hierarchical topometric map representation and an algorithm to generate it automatically from a 3D point cloud. We can generate topometric representations of varying dimensionality, suitable for a variety of applications. For example, for robotic vacuum cleaners, a 2D graph is sufficient, while for UAVs, a 3D graph is required. 0D and 1D graphs are more conducive to global navigation. The parameterized $passage$ can help ensure collision-free navigation. Comparing with state-of-the-art methods, our approach works well with different datasets.  Our algorithm has few constraints on the input point cloud; it could be multi-storey or have a slanted ceiling. 
    
    \markCM{The limitations we reviewed in Section \ref{subsec:limitation} indicate that the $column$ merge strategy, a critical component of our algorithm, still needs to be made more robust. Also, our current implementation cannot identify stairs in the point cloud because of the current $column$ merging strategy, and with stairs causing over-seg\-mentation. The problem of stairs might be solved by introducing an algebraic method, such as persistent homology in the $column$ merge strategy, to discrete slopes of the ceiling and floor. It is a direction for future work.}
    
\section{Acknowledgments}
    \label{sec:ack}
    We acknowledge the Visualization and MultiMedia Lab at University of Zurich (UZH) for the acquisition of the 3D point clouds, and colleagues in the ShanghaiTech Mobile Autonomous Robotic Systems Lab for their support to scan the rooms used we used as datasets for our evaluation.
     

	\bibliographystyle{spbasic}
	\bibliography{refs}

	
\end{document}